\pdfoutput=1
\documentclass{article}

\usepackage{PRIMEarxiv}

\usepackage[utf8]{inputenc}
\usepackage[T1]{fontenc}
\usepackage{amssymb}
\usepackage{amsmath}
\usepackage{amsthm}
\usepackage{amsfonts}
\usepackage{booktabs}
\usepackage{multirow}
\usepackage{array}
\usepackage{tabularx}
\usepackage{longtable}
\usepackage{float}
\usepackage{placeins}
\usepackage{nicefrac}
\usepackage{microtype}
\usepackage{graphicx}
\graphicspath{{media/}}
\usepackage{xcolor}
\usepackage{fancyhdr}
\usepackage[numbers,sort&compress]{natbib}
\usepackage{url}
\usepackage{hyperref}
\hypersetup{colorlinks=true, linkcolor=blue, citecolor=blue, urlcolor=blue}
\hypersetup{pdftitle={Auditing the Product, Not the Recursion: The Evaluation Evidence Behind Recursive Self-Improvement Claims},
  pdfauthor={Zihan Chen, Di Zhu, Lei Nico Zheng},
  pdfkeywords={recursive self-improvement (RSI), self-evolving agents, LLM agents, evaluation methodology, meta-research, LLM-as-a-judge}}

\theoremstyle{definition}
\newtheorem{proposition}{Proposition}

\newcommand{\FrameFreezeDate}{17 October 2026}
\newcommand{\FrameQueries}{12}
\newcommand{\FrameTotal}{173}
\newcommand{\FrameTitleHits}{85}
\newcommand{\FrameTwentySix}{153}
\newcommand{\FramePreTwentySix}{20}
\newcommand{\FrameFirstHalfTwentySix}{16}
\newcommand{\FrameSinceSep}{106}
\newcommand{\FrameMonthJul}{13}
\newcommand{\FrameMonthAug}{18}
\newcommand{\FrameMonthSep}{51}
\newcommand{\FrameMonthOct}{55}

\newcommand{\NullRecordsNinetyNine}{4.2}
\newcommand{\NullRecordsTwentyFive}{2.9}
\newcommand{\NullRecordsThousand}{6.5}
\newcommand{\NullEmaxHundred}{2.51}

\newcommand{\CAccClaimNeitherS}{14}
\newcommand{\CAccClaimPairedS}{7}
\newcommand{\CAccClaimRerunsOnlyS}{4}
\newcommand{\CAccClaimS}{45}
\newcommand{\CAccClaimSPct}{50}
\newcommand{\CAccClaimSupportedAuditPS}{6}
\newcommand{\CAccClaimSupportedS}{27}
\newcommand{\CAccClaimSupportedSPct}{60}
\newcommand{\CAccumEarlySPct}{50}
\newcommand{\CAccumLateSPct}{52}
\newcommand{\CAnchorAuditAnyPct}{95}
\newcommand{\CAnchorComparatorPct}{20}
\newcommand{\CAnchorImpClaim}{8}
\newcommand{\CAnchorImpClaimTested}{7}
\newcommand{\CAnchorInFrame}{2}
\newcommand{\CAnchorNullGateText}{20\% report a null control, while none uses a noise-aware promotion rule}
\newcommand{\CAnchorPerRoundPct}{30}
\newcommand{\CAnchorRerunsPct}{25}
\newcommand{\CAnyRerunsSPct}{26}
\newcommand{\CAuditAnyAPct}{91}
\newcommand{\CAuditAnyS}{81}
\newcommand{\CAuditAnySHi}{95}
\newcommand{\CAuditAnySLo}{82}
\newcommand{\CAuditAnySPct}{90}
\newcommand{\CAuditAnyTitlePct}{90}
\newcommand{\CAuditNoneS}{9}
\newcommand{\CAuditPPartS}{5}
\newcommand{\CAuditPReuseS}{14}
\newcommand{\CAuditPS}{27}
\newcommand{\CAuditPUnclearS}{8}
\newcommand{\CAuditYEarlySPct}{46}
\newcommand{\CAuditYLateSPct}{66}
\newcommand{\CAuditYS}{54}
\newcommand{\CAuditYSHi}{70}
\newcommand{\CAuditYSLo}{50}
\newcommand{\CAuditYSPct}{60}
\newcommand{\CBestCkptS}{11}
\newcommand{\CBothRoundS}{10}
\newcommand{\CBothRoundSPct}{11}
\newcommand{\CCodedCells}{3,420}
\newcommand{\CCodedItems}{30}
\newcommand{\CCohortAccumDiff}{2}
\newcommand{\CCohortAuditYDiff}{19}
\newcommand{\CCohortAuditYDiffHi}{$40$}
\newcommand{\CCohortAuditYDiffLo}{$-3$}
\newcommand{\CCohortMinP}{0.10}
\newcommand{\CComparatorAPct}{32}
\newcommand{\CComparatorAnySPct}{87}
\newcommand{\CComparatorEarlySPct}{27}
\newcommand{\CComparatorLateSPct}{41}
\newcommand{\CComparatorS}{33}
\newcommand{\CComparatorSHi}{47}
\newcommand{\CComparatorSLo}{27}
\newcommand{\CComparatorSPct}{37}
\newcommand{\CComparatorTitlePct}{37}
\newcommand{\CDevJudgeS}{34}
\newcommand{\CDevJudgeSPct}{38}
\newcommand{\CDevSelfS}{20}
\newcommand{\CDevSelfSPct}{22}
\newcommand{\CDivNAS}{64}
\newcommand{\CDivNotMethodA}{6}
\newcommand{\CDivTypePairedA}{11}
\newcommand{\CDivTypePartialA}{5}
\newcommand{\CDivTypeRegressionA}{7}
\newcommand{\CDivergeS}{22}
\newcommand{\CEvMutS}{14}
\newcommand{\CEvMutUS}{8}
\newcommand{\CFullProcessS}{6}
\newcommand{\CHeadDevAnySPct}{32}
\newcommand{\CHeadDevPartialS}{21}
\newcommand{\CHeadDevS}{8}
\newcommand{\CImpClaimBetterS}{22}
\newcommand{\CImpClaimNoneS}{3}
\newcommand{\CImpClaimOtherAnyS}{9}
\newcommand{\CImpClaimRelatedS}{23}
\newcommand{\CImpClaimS}{26}
\newcommand{\CImpClaimSwapMatchedS}{4}
\newcommand{\CImpClaimSwapMatchedSHi}{34}
\newcommand{\CImpClaimSwapMatchedSLo}{6}
\newcommand{\CImpClaimSwapMatchedSPct}{15}
\newcommand{\CImpClaimSwapPartialS}{3}
\newcommand{\CImpClaimTrajAnyS}{7}
\newcommand{\CNAnchor}{20}
\newcommand{\CNAnchorOutFrame}{18}
\newcommand{\CNDivergeAll}{24}
\newcommand{\CNEarlyS}{26}
\newcommand{\CNLateS}{64}
\newcommand{\CNS}{90}
\newcommand{\CNTitleS}{68}
\newcommand{\CNWeak}{24}
\newcommand{\CNullS}{17}
\newcommand{\CNullSHi}{28}
\newcommand{\CNullSLo}{12}
\newcommand{\CNullSPct}{19}
\newcommand{\COpenHumanAuditS}{1}
\newcommand{\COpenMixedGraderS}{15}
\newcommand{\COpenNoAuditS}{5}
\newcommand{\COpenPureJudgeS}{4}
\newcommand{\COpenS}{25}
\newcommand{\CPerRoundAPct}{40}
\newcommand{\CPerRoundS}{39}
\newcommand{\CPerRoundSHi}{54}
\newcommand{\CPerRoundSLo}{34}
\newcommand{\CPerRoundSPct}{43}
\newcommand{\CPerRoundTitlePct}{47}
\newcommand{\CRerunsAPct}{22}
\newcommand{\CRerunsEarlySPct}{15}
\newcommand{\CRerunsLateSPct}{25}
\newcommand{\CRerunsS}{20}
\newcommand{\CRerunsSHi}{32}
\newcommand{\CRerunsSLo}{15}
\newcommand{\CRerunsSPct}{22}
\newcommand{\CRerunsTitlePct}{22}
\newcommand{\CResultMixedS}{6}
\newcommand{\CResultNegS}{1}
\newcommand{\CScopeGapAnyS}{27}
\newcommand{\CScopeGapAnySPct}{30}
\newcommand{\CScreenEligible}{114}
\newcommand{\CSelfNS}{13}
\newcommand{\CSelfNSPct}{14}
\newcommand{\CSelfPS}{25}
\newcommand{\CSelfPSPct}{28}
\newcommand{\CSelfYS}{52}
\newcommand{\CSelfYSPct}{58}
\newcommand{\CSingleOrNRS}{66}
\newcommand{\CSingleOrNRSPct}{73}
\newcommand{\CStatGateS}{17}
\newcommand{\CStatGateSPct}{19}
\newcommand{\CSwapMatchedS}{6}
\newcommand{\CTaskLocalS}{4}
\newcommand{\CTierEFourS}{13}
\newcommand{\CTierEFourAPct}{11}
\newcommand{\CTierEFourSPct}{14}
\newcommand{\CTierFourSHi}{23}
\newcommand{\CTierFourSLo}{9}
\newcommand{\CTierFourTitlePct}{15}
\newcommand{\CUncS}{46}
\newcommand{\CUncSPct}{51}
\newcommand{\CVerifS}{65}
\newcommand{\CVerifSPct}{72}
\newcommand{\CWuConcAgree}{20}
\newcommand{\CWuConcAgreeBudget}{8}
\newcommand{\CWuConcAgreeImp}{6}
\newcommand{\CWuConcAgreeTest}{6}
\newcommand{\CWuConcCells}{24}
\newcommand{\CWuConcStricter}{2}
\newcommand{\CWuConcWorks}{8}

\newcommand{\AdjAuditAnyHi}{92}
\newcommand{\AdjAuditAnyLo}{65}
\newcommand{\AdjAuditAnyPct}{83}
\newcommand{\AdjAuditYPct}{41}
\newcommand{\AdjCells}{900}
\newcommand{\AdjComparatorPct}{34}
\newcommand{\AdjEligN}{29}
\newcommand{\AdjFirstCellPct}{81}
\newcommand{\AdjHeadDevPct}{45}
\newcommand{\AdjImpTestPct}{28}
\newcommand{\AdjNullPct}{28}
\newcommand{\AdjPerRoundPct}{45}
\newcommand{\AdjProcMinP}{0.17}
\newcommand{\AdjRerunsPct}{14}
\newcommand{\AdjRestAuditYP}{0.05}
\newcommand{\AdjRestN}{85}
\newcommand{\AdjRows}{177}
\newcommand{\AdjScopePct}{55}
\newcommand{\AdjStrongAuditYPct}{45}
\newcommand{\AdjStrongN}{22}
\newcommand{\AdjStrongRerunsPct}{14}
\newcommand{\AdjStrongScopePct}{55}
\newcommand{\AdjUpheldFirst}{3}
\newcommand{\AdjUpheldSecond}{174}
\newcommand{\AuditFourBudgetOnly}{8}
\newcommand{\AuditFourCompAgree}{72}
\newcommand{\AuditFourCompKappa}{0.50}
\newcommand{\AuditFourDesignAgree}{81}
\newcommand{\AuditFourDesignKappa}{0.74}
\newcommand{\AuditFourDisagree}{16}
\newcommand{\AuditFourN}{43}
\newcommand{\PairComparatorP}{0.73}
\newcommand{\PairPerRoundP}{1.00}
\newcommand{\PairRerunsB}{10}
\newcommand{\PairRerunsC}{2}
\newcommand{\PairRerunsP}{0.04}
\newcommand{\PairStrongRerunsP}{0.04}
\newcommand{\RelDescKappaMax}{0.91}
\newcommand{\RelDescKappaMin}{0.81}
\newcommand{\RelN}{30}
\newcommand{\RestAuditAnyPct}{94}
\newcommand{\RestAuditYPct}{64}
\newcommand{\RestHeadDevPct}{28}
\newcommand{\RestImpTestPct}{15}
\newcommand{\RestNullPct}{15}
\newcommand{\RestScopePct}{19}
\newcommand{\VOneOneScopePct}{31}

\title{Auditing the Product, Not the Recursion:\\
The Evaluation Evidence Behind Recursive Self-Improvement Claims}

\author{
  Zihan Chen \\
  Stevens Institute of Technology \\
  Hoboken, NJ, USA \\
  \texttt{zchen61@stevens.edu} \\
  \And
  Di Zhu \\
  Stevens Institute of Technology \\
  Hoboken, NJ, USA \\
  \texttt{dzhu1@stevens.edu} \\
  \And
  Lei Nico Zheng \\
  University of Massachusetts Boston \\
  Boston, MA, USA \\
  \texttt{lei.zheng@umb.edu} \\
}

\begin{document}
\maketitle

\begin{abstract}
Recursive self-improvement (RSI) promises AI systems, such as self-evolving LLM agents, that become better at producing further improvements. Evaluating this promise requires separating product claims about the final system from process claims about the improvement process itself. Existing surveys specify the evidence these claims require, but how often studies report that evidence remains unmeasured. We examine this question through full-text coding of \CScreenEligible{} eligible studies in a reproducible arXiv search frame frozen in October 2026, including \CNS{} that describe their own systems as RSI. The codebook links claims to evaluation designs. Independent human coding, third-reader adjudication, and a separate blind coding of improver comparisons assess coding reliability. Among the \CNS{} RSI studies, \CAuditAnySPct\% report product evaluation beyond development tasks under a broad criterion that includes partial or unstated separation, while \CAuditYSPct\% report evaluation on tasks unused for selection. Only \CRerunsSPct\% audit two or more reruns of the full loop. In a human-coded random sample, the RSI studies show the same contrast: \AdjStrongAuditYPct\% evaluate on tasks unused for selection and \AdjStrongRerunsPct\% audit two or more reruns. Of the \CImpClaimS{} studies claiming a better improvement procedure, \CImpClaimRelatedS{} report a related comparison, but only \CImpClaimSwapMatchedS{} compare earlier and later procedures from identical starting artifacts at a matched budget. Across these results and a comparison set of established works, evidence for repeatability and improved procedures remains less developed than product evaluation. Three propositions explain how selection effects, shared evaluator errors, and changes in system state limit what product evaluations can establish. Together, the findings support a claim-indexed reporting standard that connects each claim to the evaluation design needed to test it.
\end{abstract}


\section{Introduction}
\label{sec:intro}

Language models increasingly participate in the processes that improve AI systems. They revise prompts and agent harnesses, which coordinate a model's tools and actions \citep{zhang2025dgm,hu2025adas,xia2026rrsi,wu2026modularrsi}. They also generate feedback for updating model weights \citep{yuan2024selfrewarding,chen2024spin,yin2026opd} and modify data and training pipelines \citep{tan2026metarsi,wang2026ecgrsi}. Some systems also modify the procedure that proposes changes, which we call the \emph{improver}, or the evaluator that decides which changes to retain \citep{zelikman2024stop,zhang2026hyperagents,srikanth2026aide,wu2025metarewarding,iacob2026redqueen}. These developments motivate the ambition of recursive self-improvement (RSI): retained improvements should enable the system to produce further improvements, potentially by making the improver itself more effective.

This ambition creates an evaluation problem. A higher final benchmark score establishes something about the resulting system, but it does not by itself establish how the improvement process behaves. Whether gains accumulate across rounds, recur in independent runs, or result from a better improver are distinct empirical questions. We refer to these as \emph{process claims}, in contrast to a \emph{product claim} about the performance of the final system. Evaluating RSI requires matching the evidence to the particular claim being made.

The distinction matters because the signal used to select changes can disagree with the property the system is intended to improve. In one self-play study, a model's own judge increasingly accepts its outputs while hidden exact-match accuracy remains unchanged \citep{zhou2026convincing}. In a robotics study, accepted skill changes repeatedly pass their tests, yet the target task never succeeds \citep{wang2026robotics}. Both cases distinguish the \emph{development signal}, which guides the loop's choices, from an \emph{audit}, which assesses the claimed improvement outside those choices. Even a successful audit of the final product leaves open whether the loop produces gains repeatably or whether its improver has become better.

Recent surveys provide principles for answering these questions. \citet{wu2026rsisurvey} distinguish task gains, retention and transfer, and improver gains. Their evaluation protocol includes protected test tasks, repeated runs, resource-matched controls, and comparisons of earlier and later improvers from common starting states. \citet{wang2026reliable} require audit evidence appropriate to the component being changed, while \citet{chen2026rsi} organize feedback sources by their verification strength. These accounts establish what evidence to seek. However, they do not measure how often studies report it. The remaining question is therefore empirical: \emph{how closely does the reported evaluation of RSI systems match the claims made about their improvement processes?}

We address this question through a census of eligible studies in a reproducible arXiv search frame. We code each study from its full text, using a codebook that translates existing evaluation principles into observable features of reported experiments. The dataset covers \CScreenEligible{} eligible studies, of which \CNS{} describe their own loops or systems as RSI, and records each study's claims, development signal, evaluation boundary, grader, and process controls, with source locators identifying the supporting passages.

We ask three research questions. RQ1 asks, for each claim that RSI studies make, how often they report an evaluation design that bears on it. RQ2 asks whether the studies that claim a better improver report comparisons that can isolate an improver gain. RQ3 asks whether the findings hold under independent human coding and in established self-improvement research outside the search frame. We assess robustness through a random sample coded independently by a human and adjudicated by a third reader, separate blind coding of the improver comparisons, and a comparison set of \CNAnchor{} representative works selected by \citet{wu2026rsisurvey}, which we call anchor works.

For RQ1, the broad product-evaluation indicator covers \CAuditAnySPct\% of the \CNS{} studies. This indicator includes partial or unstated separation from development. However, only \CAuditYSPct\% report evaluation on tasks unused for any selection. Most studies that claim accumulating gains audit across rounds, whereas repeated evaluation of the loop itself is uncommon: \CRerunsSPct\% audit two or more reruns.

For RQ2, \CImpClaimRelatedS{} of the \CImpClaimS{} studies claiming a better improver report a related comparison. These comparisons usually favor the proposed process. However, only \CImpClaimSwapMatchedS{} compare an earlier and a later improver from identical starting artifacts at a matched budget. This design isolates the improver's contribution from changes in the system on which it acts. For RQ3, the human-coded sample confirms that audited reruns are rarer than selection-independent product evaluation within the same studies (\AdjStrongRerunsPct\% against \AdjStrongAuditYPct\% for its RSI studies), and the anchor works show the same pattern. Studies' own experiments further document cases in which development gains transfer only partly, or not at all, to external evaluation.

The study contributes both measurement and guidance for evaluation. First, the dataset records claims and evidence for every eligible study in the frozen search frame, including the adjudicated sample and independently coded improver comparisons. Second, the accompanying codebook makes the evaluation requirements measurable and provides evidence on coding reliability. These two resources answer RQ1--RQ3. Third, three propositions explain the limits of inferring process gains from product evidence. Finally, a reporting standard links each claim to the required evaluation, using the observed rates as a baseline.

Section~\ref{sec:related} positions the study relative to prior surveys and meta-research on evaluation practice. Section~\ref{sec:claims} defines the measured constructs, and Section~\ref{sec:model} explains why product evidence cannot stand in for process evidence. Section~\ref{sec:design} describes the data and methods, Section~\ref{sec:census} reports the results, and Section~\ref{sec:within} presents a case analysis of studies' own experiments. Section~\ref{sec:standard} discusses the findings and their implications for reporting, and Section~\ref{sec:limitations} discusses limitations and future work.

\section{Related Work}
\label{sec:related}

Our study draws on three lines of research: surveys that organize self-improving systems, research on evaluation reliability, and empirical meta-research that measures evaluation practice across published papers. The surveys identify which components can change and what those changes are intended to achieve. Evaluation research explains why an apparent gain may depend on the feedback source, the grader, or the use of test data. Meta-research supplies the method: coding published papers to measure how claims are evaluated. We apply that method to the claims of RSI research.

Surveys of self-evolving agents organize methods by the object, timing, mechanism, and setting of an update \citep{gao2026survey,fang2025comprehensive}. Related accounts describe the cycle from experience acquisition to evaluation \citep{tao2024selfevolution}, distinguish foundation-model and scaffold updates \citep{ren2026self}, and connect RSI design choices to classical learning mechanisms \citep{zhou2026twoperspectives}. Feedback taxonomies distinguish internal, external, multi-agent, and human signals \citep{liu2025feedback}, while safety surveys examine how risks persist through retained state \citep{chen2026safety,gong2026evolutionary}. Roadmaps and formal frameworks further characterize the requirements and limits of RSI \citep{duan2026last,tang2026gai,lu2026verification}.

Three surveys provide the closest basis for our coding framework. \citet{wu2026rsisurvey} distinguish task gains, retention and transfer, and improver gains. They propose a paired probe that applies earlier and later improvers to the same starting states under the same budget, alongside sealed test tasks, scheduled checkpoints, repeated runs, and fixed-improver controls. Their protocol descriptions cover eight selected works. \citet{wang2026reliable} organize self-evolution into five levels, from outputs to evaluation criteria, and match each level to external audits and controls. \citet{chen2026rsi} map 1,250 arXiv records by the component being improved and the extent of autonomous loop closure, using title and abstract keyword rules. Their verification hierarchy characterizes feedback quality qualitatively.

Two further surveys offer complementary qualitative assessments. \citet{ye2026aiforai} separate task-level refinement, lasting self-improvement and RSI, which also requires a better capacity to make further improvements, and call for independent evaluation and full cost reporting. \citet{wu2026ai4ai} synthesize evidence on AI improving AI and conclude that evidence for persistent and compounding improvement remains limited. Table~\ref{tab:positioning} places our study alongside these accounts: we operationalize their requirements through full-text coding on a defined search frame, with a separate comparison set and checks of coding agreement.

\begin{table}[t]
\caption{What the closest surveys provide and what this paper adds. ``Frame'' means a sampling frame defined independently of the authors' reading list, so that counts describe a population. A blank cell means the survey does not report the item.}
\label{tab:positioning}
\centering
\small
\begin{tabularx}{\textwidth}{>{\raggedright\arraybackslash}p{2.6cm}>{\raggedright\arraybackslash}p{1.9cm}*{4}{>{\raggedright\arraybackslash}X}}
\toprule
Work & Taxonomy of what changes & Evidence requirements & Study-level coding & Counts on a frame & Product vs.\ process \\
\midrule
\citet{wu2026rsisurvey} & six update objects & protocol: sealed tests, frozen-improver control, no-update baseline, paired improver probe & protocol notes for 8 selected works & none; prevalence disclaimed & task gain, retention, improver gain \\
\citet{wang2026reliable} & levels L0--L4 & matched-audit principle; reliability ladder & & & partly, by level \\
\citet{chen2026rsi} & substrate by loop closure & verification hierarchy (qualitative) & abstract-level keyword rules & 1,250 records (abstract-level map) & \\
\citet{ren2026self}, \citet{gao2026survey} & yes & & & & \\
\citet{ye2026aiforai}, \citet{wu2026ai4ai} & yes & research priorities; qualitative assessment of evidence & & & qualitative \\
\citet{chen2026safety} & transitions & longitudinal safety evidence & paper cards & coded pool (scoping) & \\
This paper & uses existing & existing requirements as codable items, with the design that identifies each claim & full text, every eligible study, with locators & \FrameTotal{} records, \CScreenEligible{} eligible studies; \CNAnchor{} anchor works & yes, by claim \\
\bottomrule
\end{tabularx}
\end{table}

The need to distinguish development feedback from evaluation is already apparent in task-local self-correction. Self-feedback can improve an answer through iterative refinement \citep{madaan2023selfrefine}, but reasoning corrections often fail without external signals \citep{huang2024cannot}. Tool-based critique can provide such signals \citep{gou2024critic}. Surveys synthesize these findings across correction strategies \citep{kamoi2024tacl,pan2024tacl}. Persistent self-improvement extends the question beyond a single answer. Systems retain updates based on correctness-filtered rationales \citep{zelikman2022star}, self-generated judgments \citep{yuan2024selfrewarding,wu2025metarewarding}, human-anchored self-play \citep{chen2024spin}, or AI feedback \citep{bai2022constitutional}. Agents can also preserve reflections, skills, and experience across tasks \citep{shinn2023reflexion,wang2023voyager,zhao2023expel}. Retention makes gains reusable, but it can also preserve errors: recursive data generation and self-play can narrow distributions or degrade performance \citep{shumailov2024nature,alemohammad2024mad,lee2026ocnr}.

The evaluator itself provides another source of error. LLM judges exhibit position, verbosity, and self-enhancement biases \citep{zheng2023judging,wang2024fair}. Reward models can be assessed against preference data \citep{lambert2025rewardbench}, yet continued optimization of their scores can reduce the underlying reward \citep{gao2023scaling}. Formal work characterizes reward gaming \citep{skalse2022gaming}, and experiments show that behavior learned in gameable environments can extend to reward tampering \citep{denison2024sycophancy}. Weak supervisors can nevertheless elicit stronger behavior under some conditions \citep{burns2024weak}. These results make the validity of the feedback source an empirical question, especially when an RSI loop can modify its own evaluator.

Even a fixed evaluator can become unreliable when its data are reused for selection. The reusable holdout \citep{dwork2015reusable} and the Ladder mechanism \citep{blum2015ladder} address this adaptive reuse, although empirical studies of ImageNet and Kaggle report less overfitting than worst-case theory predicts \citep{recht2019imagenet,roelofs2019meta}. \citet{sun2026reuse} apply this reasoning to RSI through REUSE, a promotion framework with a specified error guarantee. In their live experiments, the share of false promotions falls from up to 20.7\% under existing evaluation frameworks to zero. This line of work motivates the distinction between a development score used for selection and an audit used to assess the resulting claim.

Recent benchmarks make these evaluation questions concrete. They examine retained memory and skills \citep{xue2026pastbench}, data-centric research \citep{meng2026rsibench}, algorithm design \citep{chi2026ai4ai}, and embodied improvement under matched source conditions \citep{song2026mcevobench}. Other environments support RSI experiments directly \citep{meng2026rsigym,wang2026rsiforge}, track skill evolution across rounds \citep{liu2026rethinking}, or compare evolving and non-evolving agents to attribute safety failures to earlier updates \citep{das2026seabench}. Together, these studies supply the evaluation principles and controls that our census measures.

Methodologically, the study belongs to empirical meta-research on machine learning evaluation, which codes published papers or reruns published experiments to measure how results are established. \citet{gundersen2018reproducibility} scored 400 AAAI and IJCAI papers on the variables needed to reproduce them and found that none documented all of them. \citet{marie2021credibility} annotated the evaluations in 769 machine translation papers and found a growing reliance on differences in BLEU scores without significance testing or human evaluation. \citet{henderson2018deeprl} showed that variance across runs makes reported deep reinforcement learning results hard to interpret, and \citet{agarwal2021precipice} showed that most results on these benchmarks are point estimates from few runs. \citet{dodge2019showyourwork} found model comparisons whose conclusions would change with more or less computation, and \citet{kapoor2023leakage} documented leakage between training and evaluation data across many fields that adopted machine learning. These studies examine how researchers support claims about final models. RSI research adds claims about the improvement process, so we code the evidence for each claim separately. The next section specifies how the evaluation principles map to the claims made by RSI studies.

\section{Measurement Framework}
\label{sec:claims}

The codebook measures three aspects of an RSI study: the kinds of change a loop makes, the roles of development and audit evidence, and the claims that RSI studies make. We express the distinctions of \citet{wu2026rsisurvey} and \citet{wang2026reliable} as items that can be coded from a paper, with separate attention to the repeatability of the loop and the independence of the audit grader.

The first distinction concerns what changes. Following prior work \citep{kamoi2024tacl,wang2026reliable}, we separate four kinds of change. A task-local correction changes the answer or trajectory for the current task only. A persistent change alters retained state, such as weights, prompts, memory, tools, skills or harness code, and that state affects later tasks. A structural change alters the improver. A criterion change alters the evaluator or the definition of success. A study can make several of these changes at once. Retaining state does not establish that the state improves later performance. Similarly, an editable improver is not necessarily a more effective one.

Evaluating these changes requires separating two roles for feedback. The development signal guides proposals, rankings, or acceptance decisions within the loop. The signal can be a unit test, an environment outcome, a reference answer, the system's own judgment, or a learned scorer such as a reward model or an LLM judge. We use the term \emph{model judge} for such learned scorers.

An audit evaluates the claimed improvement independently of the choices made during development. This independence has two components. The first concerns the tasks. The matched-audit principle of \citet{wang2026reliable} and the sealed test tasks of \citet{wu2026rsisurvey} require that audit tasks were never used to select changes, checkpoints or hyperparameters. We call such tasks held out. The second concerns the grader: the audit grader must not share the development judge's systematic errors. A held-out benchmark scored by the same judge family that guided the search has the first kind of independence and may lack the second.

Table~\ref{tab:ladder} lists the claims that RSI papers make, from product performance to improvement of the process, with the evidence each needs and the census items that record it. The product claim (C1) says that the final system is better than the initial system under a fixed evaluation budget. It corresponds to the task gain of \citet{wu2026rsisurvey}. The other four claims concern the process. The claims are distinct rather than nested. Repeatable final gains need not rise at every round, and a later improver can outperform an earlier one at a common state even when the historical trajectory does not accumulate gains. A study therefore needs the evidence for each claim it makes, and a study that claims only a better product needs only the evidence for C1.

Accumulation (C2) says that audited performance rises across retained rounds, which scheduled audits of intermediate checkpoints can show. Repeatability (C3) says that the loop, rerun from the same start, produces gains, so the result is not one lucky trajectory. Accumulation in one run and repeatability across runs therefore require different observations.

Improver gain (C4) says that a later improver produces better successors than an earlier one from the same start and with the same budget. A \emph{swap test} checks this claim by applying the earlier and later improvers to the same starting system and auditing the successors they produce. By contrast, the frozen-improver control of \citet{zhang2025dgm} compares a run in which the improver can change with one in which it remains fixed. This evolving-versus-frozen comparison tests whether allowing the improver to change helps over a full run.

Compounding (C5) extends C4 by claiming that improver gains themselves grow over time. C4 captures the distinctively recursive claim that the improvement procedure itself becomes more effective. A loop with a fixed improver can make C1 through C3 true, and many useful systems are of this kind. Repeatability matters for any iterative loop, recursive or not.

\begin{table}[t]
\caption{Claims and their evaluation requirements. The rows distinguish product performance, accumulation, repeatability, improver gains, and compounding. The census items are introduced in Section~\ref{sec:design} and defined in Appendix~\ref{app:codebook}. The last column points to the proposition that explains why weaker evidence does not suffice.}
\label{tab:ladder}
\centering
\small
\begin{tabularx}{\textwidth}{p{0.45cm}>{\raggedright\arraybackslash}p{1.9cm}>{\raggedright\arraybackslash}X>{\raggedright\arraybackslash}p{3.6cm}>{\raggedright\arraybackslash}p{1.6cm}p{0.75cm}}
\toprule
& Claim & Evidence that licenses it & Evidence that does not & Census items & Prop. \\
\midrule
C1 & Product & Audited gain of final over initial system on held-out tasks, grader independent of the development judge, matched evaluation budget & Gain on the development signal; count of accepted changes; held-out tasks scored by a coupled judge & A1--A3; A4--A6 for attribution & \ref{prop:null}, \ref{prop:coupled} \\
C2 & Accumulation & Audited score at each retained round, on a fixed audit set & Development scores per round; audit only at the end & P1, P2 & \ref{prop:null} \\
C3 & Repeatability & Two or more independent reruns of the full loop, audited, with variation across runs and a stated target quantity (estimand) & Variation across evaluation seeds of one evolved system & P3, P3a & \\
C4 & Improver gain & Swap test: earlier and later improvers from the same start state with the same budget, successors audited & Better final product; longer trajectories; ability to edit the improver; an evolving-versus-frozen comparison alone & P4, same-start design & \ref{prop:diagonal} \\
C5 & Compounding & Repeated swap tests showing growing improver gains over a stated horizon, with total cost & Accelerating development scores; a few successful rounds & P4, repeated & \ref{prop:diagonal} \\
\bottomrule
\end{tabularx}
\end{table}

These claims also depend on the evaluation scope and comparator. The scope is the set of tasks, domains and evaluation conditions on which the claim is tested. A claim stated for open-ended domains but tested on executable benchmarks has a scope gap, even if every reported number is sound.

The comparator measures the gain obtainable with the same computational resources without recursion, for example from best-of-$N$ sampling, from a fixed improver run for the same number of proposals, or from a larger search with the initial system \citep{wu2026rsisurvey}. We call a baseline of this kind a budget-matched comparator. This comparison distinguishes gains attributable to the update mechanism from those obtainable by spending the same resources on additional search \citep{srikanth2026aide,tan2026metarsi}. The following analysis explains why these distinctions matter: selection, grading, and the changing state of the system can each make product evidence insufficient for a process claim.

\section{Limits of Product Evaluation}
\label{sec:model}

We analyze three evaluation problems behind the claim distinctions in Table~\ref{tab:ladder}: selecting among noisy scores, sharing errors between development and audit graders, and changing the system and its improver together. For each problem, we state a proposition and connect it to the corresponding census items. The first is the winner's curse: selecting the highest noisy score can overstate performance. Work on adaptive data analysis addresses this effect \citep{dwork2015reusable,blum2015ladder}, and \citet{sun2026reuse} provide a formal treatment for RSI through a certified promotion rule. We illustrate this selection effect in terms of accepted changes and reported development gain. To our knowledge, the second and third have not been stated in this form for RSI. Proofs are in Appendix~\ref{app:proofs}.

\paragraph{Setup.}
A loop runs for rounds $t=0,1,\dots$. At each round it holds a retained system $S_t$, an improver $I_t$ and a development judge $J_t$. The improver proposes candidates. The judge gives each candidate $c$ a development score $r(c)$, and a promotion rule decides whether $c$ replaces the incumbent. Let $u(s)$ denote the quantity the claim is about: the expected score of system $s$ on the reference task distribution under a fixed evaluation budget.

We write the development score as $r(c)=u(c)+e(c)$, where $e(c)$ is the development error. An audit produces $\hat a(c)=u(c)+\alpha(c)$, where the audit error $\alpha(c)$ depends on the audit tasks and grader specified below. We assume the development score, the audit score and $u$ share a scale, or can be put on one by a common affine calibration under which the additive error model holds. A merely monotone calibration preserves rankings but not expected differences, so extending the results to that case would require a separate argument. Without a common calibration, the gap between development and audited gains is not identified.

\subsection{A null-loop benchmark}

\begin{proposition}[Null-loop benchmark]
\label{prop:null}
Suppose the loop scores the initial system $s_0$ and $N$ proposals on a fixed development set, and accepts a proposal whenever its development score exceeds the incumbent's. Suppose no proposal is better: $u(c)=u(s_0)$ for every candidate. Suppose the development errors $e(s_0),e(c_1),\dots,e(c_N)$ are independent and identically distributed with a continuous distribution and variance $\sigma^2$, and that the audit error has conditional mean zero given the system, $\mathbb{E}[\alpha(S)\mid S]=0$. Then:
\begin{enumerate}
\item[(i)] the expected number of accepted proposals is $\sum_{i=2}^{N+1} 1/i$, which equals \NullRecordsTwentyFive{} for $N=25$, \NullRecordsNinetyNine{} for $N=99$ and \NullRecordsThousand{} for $N=1000$;
\item[(ii)] the reported development gain $r(S_{\mathrm{final}})-r(s_0)$ is nonnegative, and with Gaussian errors its expectation is $\sigma\, m_{N+1}$, where $m_n$ is the expected maximum of $n$ standard normal variables (for example $m_{100}=\NullEmaxHundred$);
\item[(iii)] the audited gain $\hat a(S_{\mathrm{final}})-\hat a(s_0)$ has expected value zero.
\end{enumerate}
\end{proposition}

Under this null model, development scores rise and changes are accepted even though true performance remains constant. The spurious gain grows with the number of candidates scored, not with the number of rounds. Three evaluation designs address this selection effect. An audit on tasks unused for selection, checkpoint choice, or hyperparameter tuning provides an external assessment (census item A2). A null control runs the same loop with shuffled feedback or proposals that cannot help, directly estimating the spurious gain (A5). A noise-aware promotion rule accounts for uncertainty when accepting a change (A6).

These promotion rules offer different guarantees. Repeated evaluation reduces noise in each comparison, whereas a certified rule such as REUSE also controls the error that accumulates when the same data are reused adaptively across many comparisons \citep{sun2026reuse}. The census item records both and does not distinguish them. The proposition also explains why reusing the final benchmark to choose a checkpoint weakens an otherwise held-out result, since each reuse is one more draw from which the maximum is taken.

The null-loop benchmark isolates the apparent gain that selection produces under independent development errors. It is a reference point rather than a bound for real loops, in which errors on shared development tasks are correlated, adaptively generated proposals need not have exchangeable errors, and acceptance margins change the number of records. Perfectly correlated errors, for example, produce no accepted changes at all. The conditional-mean assumption behind (iii) is what an audit with tasks and a grader that played no part in selection is meant to secure, but separation alone does not guarantee it.

\subsection{Coupled audits inherit development bias}

Separating audit tasks from selection addresses one source of optimism. A second remains when the audit grader shares systematic errors with the development judge. Write the development error as $e(c)=b(c)+\varepsilon(c)$, where $b$ is a systematic bias of the judge, such as a preference for plausible, long or self-generated outputs, and $\varepsilon$ is noise. Let the audit error be $\alpha(c)=\lambda\, b(c)+\xi(c)$. The coefficient $\lambda$ measures how strongly the audit grader shares the development judge's bias: $\lambda=0$ for a grader whose errors are unrelated to $b$, and $\lambda=1$ for the same judge used in the same way. Grader family alone does not fix $\lambda$: judges from different families can share a bias, and the same model can share less of it under a different prompt or task.

\begin{proposition}[Coupled audits]
\label{prop:coupled}
Suppose the residual audit error has conditional mean zero, $\mathbb{E}[\xi(S)\mid S]=0$, for every system $S$ the loop can select. Then the expected audited gain differs from the expected true gain by exactly $\lambda\,\mathbb{E}[\,b(S_{\mathrm{final}})-b(s_0)\,]$. If in addition $\lambda\ge 0$, each accepted change raises $b$ by at least $\delta>0$ and no change lowers it, the audited gain overstates the true gain by at least $\lambda\,\delta$ times the expected number of accepted changes.
\end{proposition}

The identity makes the conditions for an informative audit explicit. Held-out tasks make the residual condition plausible but do not guarantee it, and the sign of the distortion depends on $\lambda$ and on how $b$ changes. Two consequences follow. First, a held-out benchmark is not enough when its grader shares the development judge's bias (census item A3). The self-play judge in \citet{zhou2026convincing} learned false positives that other judge families also accepted, which is a case of $\lambda>0$ across families. Second, $\lambda$ can only be estimated against a reference grader whose errors are unrelated to $b$, such as an executable check of the claimed property, an expert panel with measured agreement, or a fixed labeled set. Verifiable tasks supply such a reference for the property they test. Open-ended tasks often do not, and this is where the independence requirement moves from the training signal to the audit.

\paragraph{Remark: criterion drift.}
When the loop changes its own judge, $J_t$ varies across rounds, and development scores from different rounds are no longer on one scale. A rising development curve can then come from the change in the criterion alone. This adds a temporal source of bias to the shared-error problem in Proposition~\ref{prop:coupled}. A fixed external reference that gates each change of evaluator, as in \citet{iacob2026redqueen}, restores comparability for the evaluators, but not for the outputs those evaluators grade.

\subsection{The diagonal problem}

The preceding results concern how gains are measured. Improver claims introduce a further question: whether gains arise from a better improvement procedure or from the state on which that procedure acts. Here $S_t$ denotes the solver state: everything in the retained system other than the improver $I_t$. We assume that any improver can be recombined with any solver state, as when the improver is a separable module.

A transition kernel $K(\cdot\mid S_t,I_t)$ specifies the probability distribution of each round's outcome: the next solver state and audited gain $(S_{t+1},\Delta_t)$. Let $g(s,I)$ be the expected audited gain under $K$ when improver $I$ acts on solver state $s$ with budget $B$. Improver gain (C4) at state $s$ means $g(s,I_{t'})>g(s,I_t)$ for a later improver $I_{t'}$. A run evaluates $K$ only at the pairs $(S_t,I_t)$ it visits.

\begin{proposition}[Diagonal non-identification]
\label{prop:diagonal}
Suppose the improver is a deterministic function of the solver state, $I_t=\phi(S_t)$, where the solver state includes whatever part of the run's history determines the improver. Then the joint distribution of states and audited gains, across any number of runs started from any initial states, depends on $K$ only through its restriction $K(\cdot\mid s,\phi(s))$. In particular, the kernel $\tilde K(\cdot\mid s,I)=K(\cdot\mid s,\phi(s))$, under which the improver has no effect on outcomes, generates the same distribution of observables. The improver contrast $g(s,I')-g(s,\phi(s))$ for $I'\neq\phi(s)$ is therefore not identified.
\end{proposition}

The proposition describes a deterministic coupling with no overlap: at each solver state the run observes only one improver. Here, non-identification means that different explanations of the improver's effect are compatible with exactly the same observed data. Co-evolution does not by itself imply this coupling, since different histories or random updates can produce different improvers at the same solver state, but loops that edit their improver only through their own retained changes come close to it. The identification problem arises because state and improver change together. Gains can shrink as the solver approaches its performance ceiling even while the improver gets better. They can also grow when a fixed improver reaches states that are easier to improve. The observed trajectory alone cannot distinguish these explanations.

A two-state example shows why matching scores does not help. Let solver states $A$ and $B$ both score zero, and let any improver produce an expected gain of 0.1 from $A$ and 0.8 from $B$. An earlier improver observed at $A$ and a later improver observed at $B$ differ by 0.7, although the improver contrast is zero at each state.

Under the unrestricted model, identification therefore requires observations off the diagonal: applying an improver to a state with which it was not originally observed. The most direct design runs an earlier and a later improver from identical starting artifacts, or from a declared distribution of starting states with paired or randomized assignment, under the same budget, and audits their successors. This is the swap test, which \citet{wu2026rsisurvey} call a paired probe and \citet{srikanth2026aide} an ignition test. It is related to clade-level metaproductivity, the performance of an agent's descendants, which \citet{wang2025hgm} use to guide search. Randomized assignment of improvers to solver states, or structural assumptions on $K$, can also supply off-diagonal information. Additional trajectory data alone cannot.

The target quantity, or estimand, should be stated: the difference in expected audited gain at one solver state, or its average over a declared set of starting states. Compounding claims extend the swap test over a stated horizon, since a few rounds of rising product scores do not establish growing improver gains.

An evolving-versus-frozen comparison at matched budget, in which one run's improver is held fixed \citep{zhang2025dgm}, answers a different question. Its frozen arm generally visits different solver states, so its total effect measures the value of letting the improver change over a run. Without shared states or further assumptions, it need not provide the overlap required for the contrast between an earlier and a later improver at a common state. These distinctions determine which comparisons the census records. Section~\ref{sec:design} describes how we translate them into coding decisions.

\section{Data and Methods}
\label{sec:design}

We answer the research questions with a quantitative content analysis of published studies. We first define the search frame and eligibility criteria, then describe the coded items, reliability checks, and analysis.

\subsection{Search frame and eligibility}

The search frame consists of arXiv records whose title or abstract contains \emph{recursive self-improvement} or a close variant, submitted up to \FrameFreezeDate. We ran \FrameQueries{} title and abstract queries through the arXiv API and took their union, obtaining \FrameTotal{} records, including \FrameTitleHits{} with the phrase in the title. Of these records, \FrameTwentySix{} were posted in 2026 and \FrameSinceSep{} on or after 1 September 2026. Appendix~\ref{app:search} provides the queries and posting chronology. Searching by label defines a reproducible population for examining how RSI claims are evaluated.

A record is eligible if it presents an implemented loop in which an AI system's outputs, evaluations, or experience are used to propose and retain changes, reports an experiment evaluating the result, and makes an improvement claim. We exclude surveys, position papers, purely theoretical analyses, and unrelated uses of the phrase. Benchmark, environment, and evaluation papers are eligible only if they also present and evaluate an improving system of their own. Records that invoke RSI solely as motivation for a method that does not improve itself are excluded, as are reports that specify a loop and an evaluation protocol but never run the loop. A study that runs its loop and describes the outcome qualitatively is eligible and enters the lowest evidence tier. Eligibility does not depend on the direction of the result, so studies whose loops fail remain in the census.

We screened titles and abstracts first, then read the full text of records that passed this stage or remained uncertain. For records with several versions, we used the latest version available on the freeze date.

Table~\ref{tab:flow} summarizes the screening process. Of the \CScreenEligible{} eligible studies, \CNS{} call their own loop or system RSI in the title, abstract, or introduction. We refer to these as \emph{strong-label studies}. The remaining \CNWeak{} use RSI as motivation, for example by describing their work as a step toward it, and are termed \emph{weak-label studies}. The primary analysis concerns the strong-label studies. We also report results for all eligible studies and for the subset with the phrase in the title.

\begin{table}[t]
\caption{Screening flow for the arXiv RSI frame.}
\label{tab:flow}
\centering
\small
\begin{tabular}{lr}
\toprule
Stage & Records \\
\midrule
Records in the frame & 173 \\
\quad Excluded on title and abstract: theory or formal model only & 16 \\
\quad Excluded on title and abstract: benchmark or environment & 11 \\
\quad Excluded on title and abstract: survey, roadmap or perspective & 7 \\
\quad Excluded on title and abstract: position paper or design proposal & 5 \\
\quad Excluded on title and abstract: analysis or evaluation study & 5 \\
\quad Excluded on title and abstract: unrelated use of the phrase & 4 \\
\quad Excluded on title and abstract: no evaluated improvement loop & 1 \\
Full texts assessed & 124 \\
\quad Excluded on full text: position paper or design proposal & 3 \\
\quad Excluded on full text: survey, roadmap or perspective & 2 \\
\quad Excluded on full text: analysis or evaluation study & 2 \\
\quad Excluded on full text: benchmark or environment & 2 \\
\quad Excluded on full text: no evaluated improvement loop & 1 \\
Eligible studies & 114 \\
\quad of which the paper calls its own loop RSI (strong label) & 90 \\
\bottomrule
\end{tabular}

\end{table}

\subsection{Coded items and dataset}

The unit of analysis is the study. We first record the claims made in its abstract, introduction, and conclusion: improved product performance, accumulation of gains, and improvement of the improver. We then code its evaluation using the items in Appendix~\ref{app:codebook}. Descriptive items identify what changes, whether the changed artifact belongs to the system running the loop, what persists between tasks, the number of rounds, the domain, the development signal, and any evaluator changes.

The evaluation items distinguish the source of the headline result (A1), the separation of product evaluation from development and selection (A2), and the audit grader (A3). A2 = Y records evaluation tasks unused for selection, including checkpoint and hyperparameter choice. We call this \emph{selection-independent evaluation}. A2 = P covers three situations that we record separately: the evaluation tasks are also used for some selection, only part of the headline result is held out, or the paper does not state the separation. We report both the strict share (Y) and the combined share (Y or P). A3 records the type of audit grader. Additional items record a budget-matched comparator (A4), a null control with removed or randomized feedback (A5), and a noise-aware promotion rule (A6).

Process items record audits across rounds or checkpoints (P1), paired development and audit scores for those rounds (P2), independent reruns of the full loop (P3), and whether at least two reruns have audit-level outcomes (P3a).

Item P4 records whether a study reports an improver-related comparison at a matched budget (Y), a partial or unmatched one (P), or none. We distinguish three designs because they answer different questions. A same-start comparison applies an earlier and a later improver to identical starting artifacts: it is the swap test discussed in Proposition~\ref{prop:diagonal}. An evolving-versus-frozen comparison tests the value of allowing the improver to change by comparing a full loop run with one in which the improver is fixed. Other comparisons include freezing an adaptive evaluator or curriculum, or swapping between different models.

Finally, P5 records reported divergence between development and audit outcomes. It is coded NA (not applicable) when no audit is paired with development scores, because divergence cannot then be assessed. S1 records whether the scope that a study presents in its title, abstract and introduction, including statements of motivation, exceeds the setting it evaluates.

Source locators connect the coding to the relevant sections, tables, figures, or appendices. The codes record which designs each study reports, whatever their results. A control that a study does not report counts as absent, not as failed. Four logical rules link the items. For example, a study without an audit cannot report a per-round audit or a development--audit divergence (Appendix~\ref{app:codebook}).

Evidence tiers summarize which combinations of these designs a study reports. E0 denotes a demonstration without quantitative evaluation of the change, and E1 denotes a gain without evaluation beyond the development tasks. E2 requires fully or partly held-out product evaluation (A2 = Y or P). E3 additionally requires a per-round audit or audited reruns, and E4 adds an improver-related comparison at a matched budget (P4 = Y). Task-local studies are capped at E2. Because P4 pools the three improver designs, E4 serves as a summary, and the primary analysis relies on claim-specific indicators that separate the designs relevant to each claim.

We apply the same codebook to \CNAnchor{} \emph{anchor works}, the representative studies selected by \citet{wu2026rsisurvey} in their Figure 8a. This set ranges from STaR \citep{zelikman2022star} and Self-Instruct \citep{wang2023selfinstruct} to the Darwin G\"odel Machine \citep{zhang2025dgm} and Hyperagents \citep{zhang2026hyperagents}. Only \CAnchorInFrame{} belong to our search frame, so the other \CNAnchorOutFrame{} provide a comparison with work outside the phrase-based search.

Wu et al.\ also describe the evaluation protocols of eight anchor works in their Table 7. We map those descriptions to three coding dimensions, test separation, improver tests, and budget parity, and compare them with our full-text codes. An improver test in their sense is a paired probe or a frozen-improver control, so we compare it with our same-start and evolving-versus-frozen designs and count other improver-related comparisons as partial.

The resulting dataset contains \CCodedItems{} coded items for each of the \CScreenEligible{} eligible studies, \CCodedCells{} codes in all, and the same evidence items for the \CNAnchor{} anchor works. Each study carries locators to the sections, tables and figures that support its codes. The dataset also records the screening decision for each of the \FrameTotal{} records in the frame, the adjudication of the double-coded sample and the independent codes of the improver comparisons.

\subsection{Coding procedure and reliability}

We piloted the codebook on 14 studies spanning harness, weight, data, improver, and evaluator updates, then froze the first version before full coding. The pilot established that evaluation separation must account for checkpoint and hyperparameter choice and that relevant evidence must be sought in appendices as well as the main text.

An LLM-based assistant supported the first coding pass by reading each full text, extracted from the PDF of the coded version, and proposing every code with a supporting locator. The later recodes were produced with the same assistance from passages retrieved from the same texts. Human coders reviewed the double-coded sample and the improver comparisons described below. No human reviewed the other codes individually.

An independent human coder, blind to the first-pass codes, then coded a random sample of \RelN{} eligible studies. Agreement was high for the label, domain, changed artifact, improver claim, reported uncertainty, and audited reruns, with Cohen's $\kappa$ between \RelDescKappaMin{} and \RelDescKappaMax{}. Agreement varied more for other evidence items (Table~\ref{tab:agreement}). The second coding was stricter about evaluation separation and more inclusive of null controls, per-round audits, and improver tests. We used these disagreements to clarify the decision rules in codebook v1.1 and recoded all eligible and anchor studies on the affected items.

A third reader then settled each of the \AdjRows{} disagreements between the two codings from the full text. The reader saw both sets of codes without knowing their source, in an order randomized by paper, and every decision required a locator. The third reader upheld the human coding in \AdjUpheldSecond{} cases and the first pass in \AdjUpheldFirst{}, so the first pass agreed with the adjudicated codes on \AdjFirstCellPct\% of the \AdjCells{} coded cells. The AI-assisted first pass therefore erred substantially more often than the independent human coding, and we treat the adjudicated codes as the reference for the sample.

Agreement statistics refer to the two blind codings before adjudication, because the revised rules were developed after examining their disagreements. Because the double-coded studies are a random sample, their adjudicated codes also estimate each headline share without relying on the first pass. Section~\ref{sec:sensitivity} compares these estimates with the codes of the remaining studies.

The census combines adjudicated codes for the \AdjEligN{} sampled studies that remain eligible with v1.1 codes for the other \AdjRestN{} eligible studies. Two subsequent revisions apply to all studies. The first (v1.2) responded to an external review of the codes. It added the design of each improver comparison and the reason for each partial separation, enforced the logical rules between items, and excluded one study whose loop was specified but never run. These additions were coded after adjudication.

The second revision (v1.3) uses independent coding of the improver comparisons to answer RQ2. A coder who was blind to our codes assessed all \AuditFourN{} studies and anchor works that claim a better improver or had been coded with an improver-related comparison. The coder recorded the improver, whether a comparison exists and whether its budget is stated as matched, its design, its result and a locator.

Before disagreements were resolved, the two codings agreed on the design for \AuditFourDesignAgree\% of the studies ($\kappa=\AuditFourDesignKappa$) and on the comparison code for \AuditFourCompAgree\% ($\kappa=\AuditFourCompKappa$). Of the \AuditFourDisagree{} disagreements, \AuditFourBudgetOnly{} concerned only whether a budget was stated as matched (Table~\ref{tab:improveraudit}).

We adopted the independent codes and then applied one rule more strictly than either coding had: a same-start comparison requires identical starting artifacts for everything except the improver, because equal scores do not imply equal states (Section~\ref{sec:model}). This rule reclassified two comparisons that both codings had counted as same-start.

\subsection{Analysis}

We report counts with their denominators. These counts summarize the recorded codes for the finite search frame. For the principal shares we add Wilson 95\% intervals, which treat the studies as draws from the research practice that generates them. These intervals do not account for coding error or missed records, which are addressed through the robustness checks and the discussion of limitations (Sections~\ref{sec:sensitivity} and~\ref{sec:limitations}). We compare studies posted before and from 1 September 2026 using Fisher's exact test, with Newcombe 95\% intervals for the differences in shares, and examine the alternative populations defined above. Because tasks, metrics, and budgets differ across studies, the analysis compares reported evaluation practices rather than pooling performance gains.

\section{Results}
\label{sec:census}

We first examine whether reported evaluation designs address the claims made by RSI studies (RQ1), then analyze improver comparisons in detail (RQ2). The remaining results describe individual evaluation items and assess robustness to coding and population choices (RQ3). The central distinction is between widespread product evaluation and the less frequently reported designs needed to assess repeatability and improver gains. Unless stated otherwise, results refer to the \CNS{} strong-label studies.

\subsection{Claim--evidence alignment (RQ1)}
\label{sec:claimalign}

Table~\ref{tab:claims} pairs each claim in Table~\ref{tab:ladder} with the studies to which it applies. For C2 and C4 these are the studies that make the claim. C1 applies to every eligible study. We also report audited reruns and budget-matched comparators for every study as recommended practices, irrespective of its explicit claims. For each row, a strict indicator records the design closest to what the claim requires, and a broad indicator records related but weaker evidence.

\begin{table}[t]
\caption{Recorded indicators relevant to each claim, strong-label studies. ``Claimants'' are the studies that make the claim in their abstract, introduction or conclusion; C1 applies to every study because every eligible study claims a better product. The last two rows report recommended practice for every study, whatever its claims. The C4 row uses the independent coding of the improver comparisons (Section~\ref{sec:improveraudit}). Counts record reported designs, whatever their results.}
\label{tab:claims}
\centering
\small
\resizebox{\textwidth}{!}{\begin{tabular}{>{\raggedright\arraybackslash}p{2.6cm}>{\raggedright\arraybackslash}p{1.6cm}>{\raggedright\arraybackslash}p{3.9cm}c>{\raggedright\arraybackslash}p{3.9cm}c}
\toprule
Claim or practice & Studies & Strict indicator & Reported & Broad indicator & Reported \\
\midrule
C1 Product gain & all & selection-independent evaluation (A2 = Y) & 54 of 90 (60\%) & evaluation beyond the development tasks (A2 = Y or P), including 8 with separation not stated & 81 of 90 (90\%) \\
C2 Accumulation & claimants & audit across rounds or checkpoints (P1 = Y) & 27 of 45 (60\%) & per-round audit or audited reruns & 31 of 45 (69\%) \\
C4 Improver gain & claimants & same-start comparison of earlier and later improvers at a matched budget & 4 of 26 (15\%) & any improver-related comparison (P4 = Y or P) & 23 of 26 (88\%) \\
\midrule
\multicolumn{6}{l}{\emph{Recommended practice, applied to every study}} \\
C3 Repeatability of the loop & all & two or more audited reruns of the loop (P3a = Y) & 20 of 90 (22\%) & two or more runs of the loop reported (P3) & 23 of 90 (26\%) \\
Gain beyond equal resources & all & comparator at a matched budget (A4 = Y) & 33 of 90 (37\%) & comparator with a reported or comparable budget (A4 = Y or P) & 78 of 90 (87\%) \\
\bottomrule
\end{tabular}
}
\end{table}

Evaluation of the final product is widespread, but its independence is less secure than the broad rate suggests. The broad product-evaluation indicator covers \CAuditAnySPct\% of studies, while \CAuditYSPct\% report tasks unused for any selection. The difference comprises studies that reuse evaluation tasks for selection, hold out only part of the headline result, or leave the separation unstated (Section~\ref{sec:items}). Among the \CAccClaimS{} studies that claim accumulating gains, \CAccClaimSupportedSPct\% audit across rounds or checkpoints. By contrast, repeated evaluation of the full loop is uncommon: only \CRerunsSPct\% of studies audit two or more reruns, and only \CAnyRerunsSPct\% report more than one run at all. Most studies compare against some baseline (\CComparatorAnySPct\%), but only \CComparatorSPct\% match its budget to that of the loop. Improver claims show the widest gap between related and identifying evidence.

\subsection{Improver comparisons (RQ2)}
\label{sec:improveraudit}

Table~\ref{tab:improver} lists every strong-label study that claims a better improver, with the comparison that the independent coder recorded for it. Of the \CImpClaimS{} claimants, \CImpClaimRelatedS{} report an improver-related comparison, and in \CImpClaimBetterS{} of these the proposed process outperforms the comparison arm. However, the comparisons answer different questions. Only \CImpClaimSwapMatchedS{} (\CImpClaimSwapMatchedSPct\%, interval \CImpClaimSwapMatchedSLo--\CImpClaimSwapMatchedSHi) compare an earlier and a later improver from identical starting artifacts at a matched budget. MetaRSI, for example, creates two copies of one released system and improves one with the original improver and the other with the meta-updated improver. Under the same budget and sealed data splits, the updated improver adds 7.2 points, compared with 4.9 for the original \citep{tan2026metarsi}.

Another \CImpClaimSwapPartialS{} run a same-start comparison without stating a matched budget. A further \CImpClaimTrajAnyS{} compare an evolving improver with a frozen one over a run, which tests whether letting the improver change helps but not whether a later improver is better at a common state. Of the rest, \CImpClaimOtherAnyS{} make other comparisons, such as freezing an adaptive evaluator or world model or comparing with a different method, and \CImpClaimNoneS{} report none.

Proposition~\ref{prop:diagonal} explains why a same-start comparison, or another source of off-diagonal observations, is needed to identify an improver gain under the unrestricted model. Two comparisons that both codings had counted as same-start fail the strict rule: SwarmRSI's inherited research state carries solution components into the new tasks \citep{meng2026swarmrsi}, and the transfer experiment of the anchor work Hyperagents moves the task agent together with the meta agent \citep{zhang2026hyperagents}.

\begin{table}[!t]
\caption{Improver-related comparisons reported by the \CImpClaimS{} strong-label studies that claim a better improver, as coded by an independent coder blind to our codes, with the strict same-start rule applied. Design: same start, an earlier and a later improver run from identical starting artifacts; evolving vs frozen, the loop run with and without the improver held fixed; other, any other improver-related comparison. Budget: matched when the paper states equal budgets, partial otherwise. Result: how the proposed process performed relative to the comparison arm. Locator: the main evidence in the coded version.}
\label{tab:improver}
\centering
\footnotesize
\setlength{\tabcolsep}{3pt}
\resizebox{\textwidth}{!}{\begin{tabular}{>{\raggedright\arraybackslash}p{2.7cm}>{\raggedright\arraybackslash}p{6.6cm}>{\raggedright\arraybackslash}p{1.9cm}c>{\raggedright\arraybackslash}p{1.4cm}>{\raggedright\arraybackslash}p{2.4cm}}
\toprule
Study & Comparison & Design & Budget & Result & Locator \\
\midrule
\citet{zelikman2024stop} & seed versus self-improved improver on the same tasks and initial solution & Same start & matched & mixed & Sec 5.1, Fig 4 \\
\citet{tan2026metarsi} & original versus meta-updated improver from the same released system & Same start & matched & better & Sec 5.4, Fig 11c \\
\citet{yuan2026sivia} & initial versus learned updater on the same parent skill and findings & Same start & matched & better & Sec 6, Tab 3 \\
\citet{lee2026mass} & generation-0 versus generation-1 model as workflow optimizer on the same tasks & Same start & matched & better & Sec 4.2, Tab 1 \\
\citet{ishibashi2024algorithm} & generators of iterations 1 to 3 applied to the same seed model & Same start & partial & better & Sec 3.2, Tab 2, Fig 4 \\
\citet{xu2026drafting} & original versus updated improver from the same released artifact & Same start & partial & better & App D.1, Tab S12 \\
\citet{zhang2026rsiflow} & meta-agents of rounds 0 to 4 on the same initial task agent & Same start & partial & better & Sec 4.4, Tab 3 \\
\citet{zheng2026dreamrsi} & evolving versus fixed hand-designed exploration policy & Evolving vs frozen & matched & better & Sec 4, Tab 1 \\
\citet{yin2026opd} & co-evolving versus frozen privileged teacher & Evolving vs frozen & matched & better & Sec 5.6, Fig 5 \\
\citet{dong2026branches} & proposal-guidance updates on versus off & Evolving vs frozen & matched & better & Sec 4.5, Tab 3b \\
\citet{wang2026ecgrsi} & data-factory operator or skill updates versus frozen at round 1 & Evolving vs frozen & matched & better & Sec 5.5, Tab 10, 15 \\
\citet{dai2026rri} & evolving versus frozen proposer prompts & Evolving vs frozen & matched & better & App E, Tab 2 \\
\citet{wang2026metaskill} & evolving versus frozen meta-skill & Evolving vs frozen & partial & better & Sec 4.2, Tab 1 \\
\citet{ye2026hace} & trained versus untrained optimizing agent & Evolving vs frozen & partial & better & Sec 4.2, Tab 3 \\
\citet{iacob2026redqueen} & co-evolving versus frozen evaluators & Other & matched & better & Sec 5.1--5.2, Tab 1--3 \\
\citet{kim2026metan} & fixed meta-operation versus OpenEvolve at equal compute & Other & matched & better & Sec 3.1, Tab 1--2 \\
\citet{yu2026hada} & hyper-agent loop versus LLaMEA at equal tokens & Other & matched & better & App D, Fig 5 \\
\citet{cui2026rsianything} & adapted versus fixed world model & Other & matched & better & Sec 3.1, Tab 2 \\
\citet{meng2026swarmrsi} & inherited versus empty research state on new tasks & Other & matched & better & App C.2, Tab 12 \\
\citet{yin2024godelagent} & self-referential agent versus Meta Agent Search at a different cost & Other & partial & better & Sec 5.1--5.3, Tab 1 \\
\citet{kakade2026polaris} & policy repair versus the original G\"odel Agent framework & Other & partial & better & Sec 4, Tab 1--2 \\
\citet{liu2026recurse} & co-evolving versus frozen checker & Other & partial & better & Sec 5.4, Tab 3 \\
\citet{xie2026riwanav} & adapted versus frozen world model & Other & partial & better & Sec IV-B, Tab I \\
\citet{guo2026aqua} & none reported & None & -- & -- & -- \\
\citet{yang2026qiushi} & none reported & None & -- & -- & -- \\
\citet{tian2026physevo} & none reported & None & -- & -- & -- \\
\bottomrule
\end{tabular}
}
\end{table}

\subsection{Product and process items}
\label{sec:items}

The studies span several forms of self-improvement. In \CSelfYS{} studies (\CSelfYSPct\%), the modified artifact belongs to the system running the loop, such as its weights, harness, memory, or improver. Another \CSelfPS{} (\CSelfPSPct\%) modify a component that the system subsequently uses, while \CSelfNS{} (\CSelfNSPct\%) improve a separate target. Only \CTaskLocalS{} studies make exclusively task-local changes. Most operate in verifiable domains (\CVerifS{}, \CVerifSPct\%). The remaining \COpenS{} include open-ended outputs graded by humans or models.

Across these settings, \CAuditAnyS{} studies (\CAuditAnySPct\%, 95\% interval \CAuditAnySLo--\CAuditAnySHi) meet the broad product-evaluation criterion (A2 = Y or P). Of these, \CAuditYS{} (\CAuditYSPct\% of all \CNS{}, interval \CAuditYSLo--\CAuditYSHi) report selection-independent evaluation, compared with \AdjStrongAuditYPct\% of the strong-label studies in the adjudicated sample (Section~\ref{sec:sensitivity}). Among the other \CAuditPS{} studies, \CAuditPReuseS{} also use the evaluation tasks for selection, such as choosing a checkpoint or reporting the best observed score. Another \CAuditPPartS{} hold out only part of the headline result, and \CAuditPUnclearS{} do not state how the evaluation tasks relate to the loop. The remaining \CAuditNoneS{} studies report no evaluation beyond development tasks.

Headline results use the development signal wholly in \CHeadDevS{} studies and partly in \CHeadDevPartialS{}, and \CBestCkptS{} report the best checkpoint on the evaluation benchmark. Uncertainty estimates for headline results appear in \CUncS{} studies (\CUncSPct\%).

\begin{figure}[t]
\centering
\includegraphics[width=0.9\textwidth]{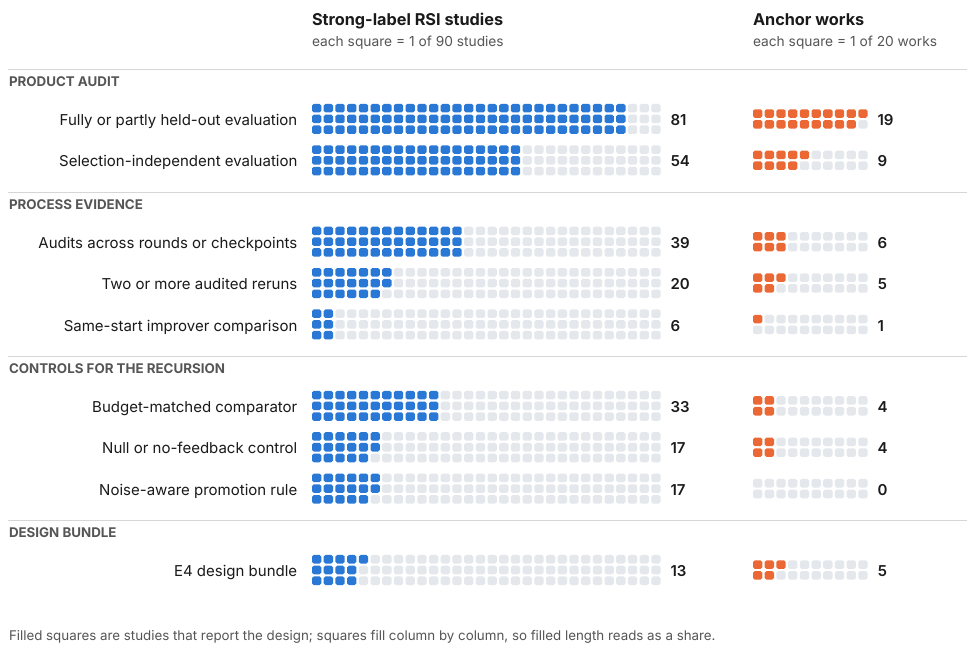}
\caption{Reported evaluation designs in the \CNS{} strong-label studies and \CNAnchor{} anchor works. The first two rows describe product evaluation, distinguishing the broad A2 = Y or P category from selection-independent evaluation (A2 = Y). The remaining rows describe process evaluation and controls. The same-start row counts every study that compares an earlier and a later improver from identical starting artifacts at a matched budget. Of the \CSwapMatchedS{} strong-label studies with such a comparison, \CImpClaimSwapMatchedS{} claim a better improver (Section~\ref{sec:improveraudit}). The E4 bundle combines fully or partly held-out evaluation, a per-round audit or audited reruns, and an improver-related comparison at a matched budget.}
\label{fig:productprocess}
\end{figure}

Figure~\ref{fig:productprocess} places these product-evaluation indicators alongside process evaluation and controls. Both forms of process evaluation are less common than the broad product indicator. Per-round audits, which include audits at scheduled checkpoints and held-out metrics of an updated component, appear in \CPerRoundS{} studies (\CPerRoundSPct\%, interval \CPerRoundSLo--\CPerRoundSHi), but only \CBothRoundS{} (\CBothRoundSPct\%) pair development and audit scores for the same rounds. Audited reruns of the full loop appear in \CRerunsS{} (\CRerunsSPct\%, interval \CRerunsSLo--\CRerunsSHi). By comparison, \CSingleOrNRS{} studies (\CSingleOrNRSPct\%) report only one loop run or do not specify the number of runs. These results distinguish evaluating the final product from assessing how gains develop across rounds or recur across runs.

The controls address a further question: what accounts for the observed gain? A budget-matched comparator appears in \CComparatorS{} studies (\CComparatorSPct\%, interval \CComparatorSLo--\CComparatorSHi). Null controls, such as shuffled feedback or random acceptance, appear in \CNullS{} (\CNullSPct\%, interval \CNullSLo--\CNullSHi), and noise-aware promotion rules in \CStatGateS{} (\CStatGateSPct\%). Only \CFullProcessS{} report all three of a per-round audit, audited reruns, and a budget-matched comparator. These counts distinguish the availability of individual checks from their use together in one study.

\subsection{Claims, scope, and evaluation graders}
\label{sec:claimsscope}

Accumulation claims deserve a closer look. Accumulation is claimed by \CAccClaimS{} studies (\CAccClaimSPct\%). Of these, \CAccClaimSupportedS{} audit across rounds or checkpoints, although \CAccClaimSupportedAuditPS{} of them rely on a partly separated evaluation. Another \CAccClaimRerunsOnlyS{} report only audited reruns, which test whether the endpoint repeats but not whether gains build over rounds, and \CAccClaimNeitherS{} report neither. Only \CAccClaimPairedS{} pair development and audit scores for the same rounds.

Table~\ref{tab:tiers} cross-tabulates the strongest claim and the evidence tier. Overall, \CTierEFourS{} studies (\CTierEFourSPct\%, interval \CTierFourSLo--\CTierFourSHi) reach the E4 bundle.

The relationship between stated and tested scope also varies. The census records scope gaps in \CScopeGapAnyS{} of the \CNS{} strong-label studies (\CScopeGapAnySPct\%). Among the \AdjStrongN{} strong-label studies in the adjudicated sample, the share is higher, at \AdjStrongScopePct\% (Section~\ref{sec:sensitivity}). This item compares the scope presented in the title, abstract, or introduction with the evaluated setting. Some authors explicitly distinguish the result they report from a stronger RSI ambition. AIDE\textsuperscript{2} reports its improver comparison as inconclusive \citep{srikanth2026aide}, and Frontis-MA1 states that it does not yet realize RSI \citep{yang2026frontis}. By contrast, Redwood describes a qualitative example as an early demonstration of RSI without quantitatively evaluating the improvement loop \citep{architect2026redwood}. These examples show why the claim and the evaluation must be coded separately.

\begin{table}[t]
\caption{Strongest reported claim and evidence tier among strong-label studies. Each study appears once, in the row of its strongest claim, so studies that claim both accumulation and a better improver appear in the improver row. E0: no quantitative evaluation. E1: no audit beyond the development tasks. E2: fully or partly held-out product evaluation. E3: E2 plus a per-round audit or audited reruns. E4: E3 plus an improver-related comparison at a matched budget. Tiers are assigned from the designs alone. The table includes \CResultNegS{} study that reports its loop did not deliver the intended improvement and \CResultMixedS{} that report mixed results.}
\label{tab:tiers}
\centering
\small
\begin{tabular}{lcccccc}
\toprule
Highest claim & E0 & E1 & E2 & E3 & E4 & Total \\
\midrule
Product only & 1 & 5 & 20 & 10 & 0 & 36 \\
Accumulation & 1 & 2 & 8 & 15 & 2 & 28 \\
Improver & 0 & 0 & 8 & 7 & 11 & 26 \\
\midrule
All strong & 2 & 7 & 36 & 32 & 13 & 90 \\
\bottomrule
\end{tabular}

\end{table}

Evaluation also depends on who supplies the scores. The development signal includes the actor's own judgment in \CDevSelfS{} studies (\CDevSelfSPct\%), and either self-judgment or a model judge in \CDevJudgeS{} (\CDevJudgeSPct\%). In \CEvMutS{} studies, the system changes its own evaluator, rubric, or tests, and \CEvMutUS{} report no fixed external reference governing those changes. These designs make grader provenance relevant to the interpretation of product and process results.

Table~\ref{tab:graders} reports evaluation graders by domain. Verifiable-domain studies predominantly use executable checks or reference labels. Among the \COpenS{} open or mixed-domain studies, the audit is graded by human raters alone in \COpenHumanAuditS{}, by a model judge alone in \COpenPureJudgeS{}, and by a combination of graders in \COpenMixedGraderS{}. The other \COpenNoAuditS{} report no audit beyond the development tasks. Mixed designs include labels with a model judge, preference scores with a user study, and panels containing the system's own evaluator. Human raters, judges from another family and executable tests can all share errors with the development signal, and Proposition~\ref{prop:coupled} shows how such shared errors bias the audited gain. Some studies introduce explicit reference checks: the Red Queen G\"odel Machine evaluates replacement graders against fixed objective or human labels \citep{iacob2026redqueen}, while RRI uses adversarial search to identify and repair hackable rubrics \citep{dai2026rri}.

\begin{table}[t]
\caption{Reported evaluation grader by domain, strong-label studies. Grader type is coded separately from evaluation separation (A2).}
\label{tab:graders}
\centering
\small
\begin{tabular}{lccc}
\toprule
Audit grader & Verifiable & Open or mixed & All strong \\
\midrule
Execution, simulator or environment & 33 & 0 & 33 \\
Reference labels & 24 & 0 & 24 \\
Human raters & 0 & 1 & 1 \\
Model judge, other family & 0 & 4 & 4 \\
Mixed graders & 4 & 15 & 19 \\
No audit recorded (A3 = NA) & 4 & 5 & 9 \\
\midrule
Total & 65 & 25 & 90 \\
\bottomrule
\end{tabular}

\end{table}

\subsection{Robustness to coding and population (RQ3)}
\label{sec:sensitivity}

RQ3 asks whether these findings depend on the coding or on the study population. The double-coded studies are a random sample of the eligible studies, so their adjudicated codes give an estimate of each headline share that does not rest on the AI-assisted first pass. Table~\ref{tab:consensus} reports these estimates for the \AdjEligN{} sampled studies that remain eligible, together with the first-pass codes for the same studies and the codes of the other \AdjRestN{} eligible studies.

In the adjudicated sample, selection-independent evaluation remains more common than audited reruns, at \AdjAuditYPct\% against \AdjRerunsPct\%: within the same studies, \PairRerunsB{} report the first without the second and \PairRerunsC{} the reverse (exact paired $p=\PairRerunsP$). The broad product indicator, at \AdjAuditAnyPct\% (interval \AdjAuditAnyLo--\AdjAuditAnyHi), exceeds every process item, while selection-independent evaluation is about as common as per-round audits at \AdjPerRoundPct\% ($p=\PairPerRoundP$) and budget-matched comparators at \AdjComparatorPct\% ($p=\PairComparatorP$). The \AdjStrongN{} strong-label studies in the sample show the same pattern, with selection-independent evaluation in \AdjStrongAuditYPct\% and audited reruns in \AdjStrongRerunsPct\% ($p=\PairStrongRerunsP$). Human coding therefore supports a gap in audited reruns. The independent improver coding identifies a separate gap in same-start comparisons (Section~\ref{sec:improveraudit}).

Comparing the adjudicated sample with the remaining studies assesses whether the headline shares are sensitive to the AI-assisted coding. For no process item does the share among the other studies differ significantly from the adjudicated share (smallest $p=\AdjProcMinP$). The largest gaps concern null controls, recorded in \RestNullPct\% of the other studies against \AdjNullPct\% of the sample, and matched improver comparisons, at \RestImpTestPct\% against \AdjImpTestPct\%. The census may therefore undercount null controls. For improver comparisons, both groups use the independent coding.

The product indicators show a consistent directional difference. Compared with the adjudicated codes, the codes outside the sample record fully or partly held-out evaluation more often (\RestAuditAnyPct\% versus \AdjAuditAnyPct\%), selection-independent evaluation more often (\RestAuditYPct\% versus \AdjAuditYPct\%, $p=\AdjRestAuditYP$), and headline results based on the development signal less often (\RestHeadDevPct\% versus \AdjHeadDevPct\%). They also record a scope gap far less often (\RestScopePct\% versus \AdjScopePct\%, $p<0.01$).

These comparisons suggest that the census overstates product evaluation, particularly its independence from selection, and understates how often the stated scope exceeds the tested one. These differences make product evidence look stronger in the census than in the adjudicated sample. Apart from the scope gap (\VOneOneScopePct\% versus \AdjScopePct\%) and matched improver comparisons, which use the independent coding, the v1.1 codes of the sampled studies match the adjudicated shares closely. However, they were revised after the second coding of these same studies, so they do not show how accurate v1.1 is elsewhere.

\begin{table}[t]
\caption{Headline items in the double-coded random sample and in the other eligible studies, strong and weak label combined. First pass: the blind AI-assisted codes. v1.1 recode: the revised codes of the same studies before adjudication, written after their second coding. Adjudicated: the codes after a third reader settled every disagreement between the first pass and the independent human coding; these enter the census. The interval is a Wilson 95\% interval for the adjudicated share. The other eligible studies carry v1.1 codes. In both groups, matched improver comparisons come from the independent coding of Section~\ref{sec:improveraudit}. Fisher's exact test compares the adjudicated sample with the other studies.}
\label{tab:consensus}
\centering
\small
\resizebox{\textwidth}{!}{\begin{tabular}{lcccccc}
\toprule
& \multicolumn{4}{c}{Double-coded random sample ($n=29$)} & Other eligible & \\
\cmidrule(lr){2-5}
Item & First pass & v1.1 recode & Adjudicated & 95\% interval & studies ($n=85$) & Fisher $p$ \\
\midrule
Fully or partly held-out evaluation (A2 = Y or P) & 93\% & 83\% & 83\% & 65--92 & 94\% & 0.12 \\
Selection-independent evaluation (A2 = Y) & 66\% & 41\% & 41\% & 26--59 & 64\% & 0.05 \\
Headline uses development signal (A1 = Y or P) & 28\% & 41\% & 45\% & 28--62 & 28\% & 0.11 \\
Per-round audit (P1 = Y) & 24\% & 45\% & 45\% & 28--62 & 39\% & 0.66 \\
Audited reruns (P3a = Y) & 17\% & 17\% & 14\% & 5--31 & 25\% & 0.30 \\
Budget-matched comparator (A4 = Y) & 38\% & 38\% & 34\% & 20--53 & 32\% & 0.82 \\
Null control (A5 = Y) & 7\% & 28\% & 28\% & 15--46 & 15\% & 0.17 \\
Noise-aware promotion (A6 = Y) & 10\% & 17\% & 14\% & 5--31 & 20\% & 0.58 \\
Matched improver comparison (P4 = Y) & 14\% & 17\% & 28\% & 15--46 & 15\% & 0.17 \\
Tier E4 & 10\% & 14\% & 14\% & 5--31 & 11\% & 0.74 \\
Scope gap (S1 = Y or P) & 17\% & 31\% & 55\% & 38--72 & 19\% & $<$0.01 \\
\bottomrule
\end{tabular}
}
\end{table}

The anchor works provide a complementary comparison (Figure~\ref{fig:productprocess}; Table~\ref{tab:anchor} in Appendix~\ref{app:codes}). Fully or partly held-out product evaluation appears in \CAnchorAuditAnyPct\% of these works, compared with \CAnchorPerRoundPct\% for per-round audits, \CAnchorRerunsPct\% for audited reruns, and \CAnchorComparatorPct\% for budget-matched comparators. In addition, \CAnchorNullGateText. Of the \CAnchorImpClaim{} works claiming a better improver, \CAnchorImpClaimTested{} report an improver-related comparison at a matched budget, but only STOP compares an earlier and a later improver from identical starting artifacts. These established works show a similar pattern beyond the RSI search frame.

For eight anchor works, we also compare our codes with the protocol descriptions in \citet{wu2026rsisurvey} (Table~\ref{tab:wuconc}). The descriptions and codes agree in \CWuConcAgree{} of \CWuConcCells{} cells: \CWuConcAgreeTest{} of \CWuConcWorks{} for test separation, \CWuConcAgreeImp{} of \CWuConcWorks{} for improver tests, and \CWuConcAgreeBudget{} of \CWuConcWorks{} for budget parity. The \CWuConcStricter{} test-separation disagreements concern ADAS's choice of transfer agents by test accuracy \citep{hu2025adas} and Meta-Rewarding's comparison of length-control settings on its reported benchmark \citep{wu2025metarewarding}. The other two disagreements concern improver tests. Wu et al.\ list no improver test for Meta-Rewarding, whereas we record its comparison with Self-Rewarding at equal prompt and iteration counts as another improver-related comparison, which counts as partial. For Promptbreeder \citep{fernando2024promptbreeder}, the independent coder records the ablation with frozen mutation prompts as matched in population and evaluations, whereas Wu et al.\ describe it as an operator ablation, which implies a partial code.

A temporal comparison separates the \CNLateS{} strong-label studies posted from 1 September 2026 from the \CNEarlyS{} posted earlier. Selection-independent evaluation appears in \CAuditYLateSPct\% versus \CAuditYEarlySPct\%; per-round audits or audited reruns in \CAccumLateSPct\% versus \CAccumEarlySPct\%; budget-matched comparators in \CComparatorLateSPct\% versus \CComparatorEarlySPct\%; and audited reruns in \CRerunsLateSPct\% versus \CRerunsEarlySPct\%. The largest difference is in selection-independent evaluation, \CCohortAuditYDiff{} percentage points in favor of the later cohort (95\% interval \CCohortAuditYDiffLo{} to \CCohortAuditYDiffHi{}).

Fisher's exact tests do not detect a difference on any of the four indicators (smallest $p=\CCohortMinP$), but the intervals are wide. All four point estimates favor the later cohort, by \CCohortAccumDiff{} to \CCohortAuditYDiff{} points, so the data are compatible both with stable practice and with a substantial improvement during the increase in RSI publications.

Finally, the shares are similar under alternative eligibility subsets. Among all \CScreenEligible{} eligible studies, \CAuditAnyAPct\% report fully or partly held-out product evaluation, \CPerRoundAPct\% a per-round audit, \CRerunsAPct\% audited reruns, \CComparatorAPct\% a budget-matched comparator, and \CTierEFourAPct\% E4. For the \CNTitleS{} strong-label studies with the phrase in the title, the corresponding shares are \CAuditAnyTitlePct\%, \CPerRoundTitlePct\%, \CRerunsTitlePct\%, \CComparatorTitlePct\%, and \CTierFourTitlePct\%. Domain differences are reported in Table~\ref{tab:matrix} in Appendix~\ref{app:codes}. The following section examines what individual studies show about the practical consequences of these evaluation distinctions.

\section{Case Analysis: What Studies' Own Experiments Show}
\label{sec:within}

The census measures how often evaluation designs are reported. This section examines what these designs show by comparing development and audit outcomes within individual studies. In \CNDivergeAll{} eligible studies, including \CDivergeS{} with the strong label, the authors report development gains that transfer only partly, or not at all, to an audit (P5 = Y or P). Table~\ref{tab:within} in Appendix~\ref{app:codes} collects these cases. They document several forms of divergence.

In \CDivTypePairedA{}, the development signal and the audit disagree about the same systems: one rises or ranks candidates while the other stays flat, falls, or ranks them differently. In \CDivTypePartialA{}, the audit confirms a smaller gain or the gain does not carry to a shifted domain. In \CDivTypeRegressionA{}, audited gains reverse in later rounds or fail to replicate across runs, and in one the audit itself was compromised. In \CDivNotMethodA{} of the cases, the divergence concerns an ablation, a baseline or an unpromoted candidate rather than the proposed method.

The availability of paired outcomes determines which cases can be examined. Most studies do not report development and audit scores in a form that would show a divergence: \CDivNAS{} of the \CNS{} strong-label studies are coded P5 = NA. We examine the cases alongside improver tests and scope comparisons, and we also discuss a few examples, such as grader dependence, that the divergence item does not count.

\subsection{Development gains and external outcomes}
The robotics loop of \citet{wang2026robotics} accepted a change only after it passed replays, paired trials on fresh scenes and a held-out task. Yet the target task never succeeded over 123 rounds. The reach evaluator used a spherical boundary, and the agent learned to stop exactly on its surface. The authors report about two human repairs to the harness for every accepted skill change.

Other studies report partial transfer or regression across rounds. In ECG-RSI \citep{wang2026ecgrsi}, the error rate on the feedback set fell from 39.63\% to 21.33\% across three rounds, while successive gains on seven external benchmarks shrank from 13.07 to 3.01 to 0.38 points. Three benchmarks regressed in the last round.

RRSI \citep{xia2026rrsi} shows the pattern across methods. Prior methods for evolving harnesses transferred little of their gain on development tasks to out-of-distribution evaluation. The arm without RRSI's regularizers reached the highest evolution score of any arm, 92.8, while its out-of-distribution average, 40.3, stayed within a point of the unevolved harness. The Mendel G\"odel Machine reports a similar reduction in gains for coding agents \citep{liu2026mendel}. Evolution raised accuracy on its 60 Polyglot evolution tasks from 50.8\% to 93.2\%, while the resulting scaffold, frozen for evaluation, gained 10.0 and 13.3 points on held-out SWE-bench Pro and SWE-bench Multilingual. The scaffold evolved by its HGM baseline lost 3.4 points on SWE-bench Pro.

A more direct disagreement between development and audit scores appears in R$^2$Flow: most edits admitted on reward-derived labels raised the training reward while lowering the held-out verified score \citep{zhang2026r2flow}. VideoHarness-SRSI documents a baseline whose validation accuracy keeps rising over rounds while accuracy on held-out questions does not \citep{lin2026videoharness}, and RecurSE describes a related effect as validation deception \citep{liu2026recurse}.

Longer runs can also show a performance peak followed by regression: DataFoundry is best after two rounds, and its Finance score falls from 66.1 to 58.3 after one more \citep{yang2026datafoundry}, while DA-RSIR peaks at round 4 \citep{hasan2026darsir}. In EvoAudio, a candidate that gains on the held-out promotion set ``does not always gain on the benchmarks'' \citep{wang2026evoaudio}. A-Evolve-Training used a public leaderboard inside its loop. Its development metric reached record highs without improving the external target, and its authors do not interpret the result as evidence of RSI \citep{shi2026aevolve}. These cases motivate measuring development and audit outcomes together rather than inferring one from the other.

Variation between runs adds another evaluation question. Proposition~\ref{prop:null} describes how selecting the best noisy score can inflate a reported gain, and two studies examine this selection effect. In Athanor, the first evaluation seed exceeded the mean of later seeds for 19 of 23 re-evaluated best candidates (sign test $p=0.003$) \citep{zhou2026athanor}. MedForge-RSI pre-registered its configurations and found that the best single-run result often failed to replicate across seeds \citep{chen2026medforge}.

Benchmarks designed to study improvement loops report related effects. In RSIBench-Data, 78.26\% of agent searches that continued past their best score ended on a lower final attempt \citep{meng2026rsibench}, and in a controlled study of skill evolution, 9 of 11 validation-selected skills improved released-test performance \citep{liu2026rethinking}. These results motivate reporting variation across full loop runs, separately from evaluation variation for one evolved system.

\subsection{Dependence on the evaluation grader}
A separate concern arises when development and audit graders share errors. In the self-play study introduced in Section~\ref{sec:intro}, the judge's pass rate rises from 0.72 to 0.94 while hidden exact-match accuracy remains at 0.20 across three seeds \citep{zhou2026convincing}. Judges from other model families and a strict three-judge ensemble also accept many of the policy's errors. This study lies outside the search frame but illustrates the shared-error mechanism in Proposition~\ref{prop:coupled}. Self-preference persists even under objective rubrics \citep{pombal2026selfpref}.

Within the frame, RSIAgent reports that its internal verifier agent returned PASS where the official rubric failed \citep{zhu2026rsiagent}. In the Red Queen G\"odel Machine \citep{iacob2026redqueen}, each change of evaluator is checked against a fixed reference, but the writers and provers are scored by panels that include the system's own co-evolved reviewer or grader. Without its own grader in the panel, the evolved prover matches a human-engineered baseline (4.03 versus 4.08) rather than exceeding it.

MASS \citep{lee2026mass} illustrates a related dependence: one model acts as agent, evaluator and optimizer. Its synthetic tasks were generated by a model that is also one of the two external judges, and its judge-graded gains are accompanied by flat results on two benchmarks with executable tests.

Other studies introduce reference checks to separate evaluation from the development judge. SpecForge \citep{cheng2026specforge} searches against tests that the model generates and evaluates on ground-truth tests hidden from the search, and MedRSI \citep{wu2026medrsi} validates its clinical-consequence judge against adjudicated references and then freezes it.

\subsection{Testing the improver and specifying scope}
Grader validation addresses how outcomes are scored, whereas improver tests address what produces them. Among self-modifying coding agents, an agent's own benchmark score can diverge from the performance of the descendants it produces \citep{wang2025hgm}, illustrating the distinction between solver performance and improvement ability. The designs distinguished in Section~\ref{sec:improveraudit} make this difference testable.

A same-start comparison runs an earlier and a later improver from identical starting artifacts. STOP measures the improver of each iteration on the same held-out tasks and initial solutions \citep{zelikman2024stop}, and Sivia applies its initial and its learned updater to the same parent skill \citep{yuan2026sivia}. The result of such a comparison depends on its conditions: STOP's self-improved improver helps with GPT-4 but not with GPT-3.5 or Mixtral.

A direct comparison can also yield an inconclusive result. AIDE\textsuperscript{2} runs a discovered agent and the human-engineered agent as outer-loop improvers from the same start, three seeds each. The arms end at 0.780 and 0.782, and the authors report the result as inconclusive \citep{srikanth2026aide}.

An evolving-versus-frozen comparison evaluates the whole loop under two conditions, allowing the improver to change in one and holding it fixed in the other, as in the Darwin G\"odel Machine \citep{zhang2025dgm} and Hyperagents \citep{zhang2026hyperagents}. It tests whether letting the improver change helps over a run, but because the two arms visit different states, it does not by itself show that a later improver beats an earlier one at a common state.

A third group of comparisons freezes another adaptive component, such as the evaluator in the Red Queen G\"odel Machine \citep{iacob2026redqueen} or a curriculum rule recomputed from the current model. These ablations test the value of adaptation rather than the improvement of the improver.

Interpreting these tests also requires matching the stated scope to the evaluated setting. HELIX names model--harness co-evolution in its title and states in its limitations that it does not train an updated model \citep{fan2026helix}, so the title describes a capability the study does not test. The MetaRSI abstract argues that RSI must eventually operate in scientific domains where correctness is settled by argument, replication or measurement, while its experiments use executable coding and closed-form science \citep{tan2026metarsi}. This statement describes the motivation for the work rather than a claim about the present system. S1 records both kinds of gap because both appear in the framing of a study. Scoping can also be explicit: Frontis-MA1 describes itself as a step toward meta-evolution that does not yet realize RSI \citep{yang2026frontis}, so its evidence for a better product matches its claim.

Together, these cases show why the development signal, audit, process controls and stated scope need to be reported as one account of the experiment. The discussion translates these evaluation needs into reporting recommendations.

\FloatBarrier
\section{Discussion}
\label{sec:standard}

\subsection{Interpreting the findings}

The findings identify two priorities for process evaluation: audited reruns and comparisons that isolate improver gains. Product evaluation is widely reported, although its separation from selection is often partial or unstated. Repeated runs of the full loop are expensive when each run trains models or executes many agent episodes, and \CSingleOrNRSPct\% of the studies report one run or do not state how many they ran. Deep reinforcement learning shows the same tendency, where computationally demanding benchmarks have led to evaluation on few runs \citep{agarwal2021precipice}.

The improver gap concerns experimental design. Most studies claiming a better improver report a related comparison, and these comparisons usually favor the proposed method. Many, however, test the value of adaptation or compare different methods without isolating the difference between an earlier and a later improver at a common state. Proposition~\ref{prop:diagonal} implies that more experiments of the same kind would not close this gap. Closing it requires a different design.

These results extend meta-research on machine learning evaluation to claims about a process. Earlier studies measured documentation, significance testing, run counts and leakage for claims about a final model \citep{gundersen2018reproducibility,marie2021credibility,kapoor2023leakage}. In RSI research the relevant unit is the claim, because a study can meet every requirement of a product claim and still report nothing that bears on the process it names. The same claim-level coding could be applied to other research whose claims concern how systems are produced, such as automated machine learning.

\subsection{Implications for reporting}

The findings and cases point to one reporting need: readers should be able to trace an improvement claim to the experiment that evaluates it. We therefore index the standard by claim (Table~\ref{tab:standard}). The general requirements apply to every study, while the remaining rows specify the evidence needed for each claim. A study claiming only a better product should therefore follow the general row and the C1 row. The recommendations build on \citet{wu2026rsisurvey} and \citet{wang2026reliable}, and the last column of the table gives the measured rate of each row's strict indicator as a baseline.

\begin{table}[t]
\caption{A claim-indexed reporting standard. Items under ``always'' apply whenever the claim is made; items under ``when relevant'' apply under the stated condition. The last column gives the share of strong-label studies, or of the studies making the claim, that report the strict indicator of Table~\ref{tab:claims}, the component of the row that the census measures.}
\label{tab:standard}
\centering
\small
\begin{tabularx}{\textwidth}{>{\raggedright\arraybackslash}p{2.1cm}>{\raggedright\arraybackslash}X>{\raggedright\arraybackslash}X>{\raggedright\arraybackslash}p{1.7cm}}
\toprule
Claim & Report always & Report when relevant & Census rate \\
\midrule
Any claim & The claim and its scope; development signal and audit, with their tasks, graders and roles; how the reported checkpoint was chosen; candidates, seeds and cost & Changes to the evaluator, rubric or tests, with scores on a fixed reference; human repairs to the system & -- \\
C1 Product & Final versus initial system on tasks unused for any selection; the audit grader's relation to the development judge & A matched-budget comparator when the gain is attributed to the loop; a null control or noise-aware promotion when many candidates are scored; validation of a model judge that grades the audit & \CAuditYSPct\% \\
C2 Accumulation & Audited scores at predeclared rounds or checkpoints, with development scores for the same rounds & The horizon over which accumulation is claimed & \CAccClaimSupportedSPct\% of claimants \\
C3 Repeatability & Two or more independent runs of the full loop from the same start, with audited outcomes, their spread and the estimand, such as the expected gain or the probability of a positive gain & The number of runs needed for the stated precision; if reruns are infeasible, a claim restricted to one trajectory & \CRerunsSPct\% \\
C4 Improver gain & An earlier and a later improver from identical starting artifacts, or from a declared set of starting states with paired assignment, at a matched budget, with successors audited and the direction and uncertainty of the contrast & Evolving-versus-frozen comparisons as supporting evidence; an inconclusive result reported as such & \CImpClaimSwapMatchedS{} of \CImpClaimS{} claimants \\
C5 Compounding & Repeated same-start comparisons over a stated horizon; the quantity claimed to grow, such as the improver's advantage or gain per unit of cost; total cost & -- & not measured \\
\bottomrule
\end{tabularx}
\end{table}

The first row asks for what every claim needs. A paper should state which claim in Table~\ref{tab:ladder} it makes, what persists between rounds, and the tasks and domains to which the result applies. This is particularly relevant to the \CScopeGapAnySPct\% of strong-label studies coded with a gap between stated and tested scope, and to the \AdjScopePct\% found in the adjudicated sample. The description of the evaluation should then state whether any reported test was used for selection, checkpoint choice or hyperparameter tuning, and should distinguish a final or prespecified checkpoint from the best checkpoint observed on a benchmark. In this census, \CAuditYSPct\% of studies report selection-independent evaluation, while \CHeadDevAnySPct\% use the development signal wholly or partly in a headline result.

The grader needs a corresponding description. When a model judge supplies audit scores, authors should state its relationship to the actor and the development judge. They should also report its agreement with a reference, such as executable checks of the same property or expert labels, on outputs produced after optimization. When the system's own evaluator sits on an audit panel, reporting results with and without it shows how much the score depends on that evaluator, as in \citet{iacob2026redqueen}. These details connect the observed score to the shared-error concern of Proposition~\ref{prop:coupled}.

Controls are conditional because they answer specific questions. A protected final test can establish a product gain without a null-loop experiment. A null control, such as shuffled feedback or inert proposals, estimates the gain that selection alone produces, which matters when many candidates are scored. A promotion rule with a stated error guarantee bounds false acceptance under its assumptions \citep{sun2026reuse}. A budget-matched comparator tests whether the update mechanism beats other uses of the same resources. The census rates are \CNullSPct\% for null controls, \CStatGateSPct\% for noise-aware promotion and \CComparatorSPct\% for budget-matched comparators.

Resource accounting answers three separate questions. Performance at an equal evaluation or inference budget concerns the product. Improvement yield at an equal search or training budget concerns the loop, and it is what a budget-matched comparator measures. End-to-end cost, including evaluator calls, data, failed candidates and human work, determines whether running the loop is worthwhile. Matching the number of proposals does not by itself match tokens, training compute or wall-clock time, so the matched quantity should be named.

Process claims need observations of the process itself. For accumulation, a predeclared audit schedule suffices for a finite-horizon claim, provided that the schedule matches the claim. For repeatability, two runs document repetition but do not by themselves establish a reliable gain. Authors should state the quantity they aim to estimate and the precision of the estimate.

For an improver gain, the comparison should apply both improvers to identical starting artifacts, or pair them over a declared set of starting states, while holding budgets fixed. Equal starting scores are not enough. Authors should also specify whether the contrast concerns one state or an average over the set. An inconclusive comparison, as in \citet{srikanth2026aide}, is a tested claim with an uncertain result, not an absent test.

Finally, changes to an evaluator, rubric or test suite should be documented alongside the system updates: what changed, why, who authorized it, and the scores before and after on a fixed reference. Human repairs to the harness belong in this record, as the robotics study of \citet{wang2026robotics} illustrates.

\section{Limitations and Future Work}
\label{sec:limitations}

\paragraph{External validity.} The census describes reported evaluation practices within a defined literature frame. Its phrase-based search captures arXiv records that use the RSI label and omits work on similar mechanisms that does not use that phrase. The anchor works provide a useful comparison largely outside the frame, but they are a selected set rather than a representative sample. Their similar pattern does not establish prevalence across self-improvement research. Search tokenization and the freeze date also affect the retrieved population. We therefore report the exact queries and chronology in Appendix~\ref{app:search}. The temporal comparisons detect no statistically significant differences, but their intervals include substantial change, so they do not establish that evaluation practice is stable.

\paragraph{Measurement reliability.} The codes describe what papers report, not all experiments their authors may have conducted. Distinguishing strong and weak labels, partial and selection-independent evaluation, or explicit and implicit scope requires judgment. An LLM-based assistant supported the first coding pass and the later recodes, and the adjudication shows that this pass erred far more often than the independent human coding. We replaced the codes of the double-coded studies with the adjudicated ones, adopted the independent codes for the improver comparisons, and revised the decision rules for the rest. Outside these two sets, individual codes have not been checked by a human. The independent improver coding covers studies with an improver claim or a previously coded comparison. A missed comparison in a study with neither would therefore go undetected. The random sample supports checks of aggregate shares, not corrections of individual codes, and it indicates that these codes credit product evaluation, especially selection-independent evaluation, too often and record scope gaps too rarely.

A single third reader settled the disagreements, so the adjudicated codes reflect one interpretation of the codebook. Adjudication produces a consensus rather than a reliability estimate. The improver comparisons received a separate blind coding pass, followed by application of the stricter same-start rule. The other final rules have not been independently applied to a fresh sample of the remaining studies. The reasons for partial separation and the types of divergence also lack a second coding.

Eligibility screening was not independently checked, so false exclusions of benchmark and evaluation papers that run an improving system of their own remain possible. The comparison with protocol descriptions from \citet{wu2026rsisurvey} checks selected interpretations rather than providing an additional reliability estimate.

\paragraph{Construct validity.} Some distinctions exceed what the current codes measure. Grader type does not establish independence of errors, and the codebook does not record reference validation for every audit grader. The unit of coding is the study, so a study can combine a held-out product test on one benchmark with an improver comparison on another. The E4 bundle therefore does not guarantee that a single experiment meets the requirements of C4. The scope item compares the framing in the title, abstract and introduction with the tested setting, so it records gaps in framing, including statements of motivation, rather than a rate of overclaiming.

The per-round item includes scheduled checkpoints and held-out metrics of an updated component, so it is broader than an audit at every round. Evidence tiers summarize reported designs rather than successful demonstrations, and the divergence cases are documented examples of several kinds, not a failure rate. Several claim-conditioned populations are small, including the \CImpClaimS{} improver claimants, so their shares carry wide intervals. These boundaries are important when moving from descriptive counts to judgments about individual methods.

\paragraph{Scope of the propositions.} The propositions isolate evaluation problems under stylized assumptions. The null model assumes a common affine score scale, independent and identically distributed development errors, and conditionally unbiased audit errors. The other propositions use a linear shared-bias model and a deterministic relationship between a separable solver state and its improver, respectively. They explain limits of inference under those assumptions rather than estimating the size of an effect in a particular study. The paper therefore does not establish that current loops fail to improve systems. Its contribution is to distinguish the evidence reported for products from that reported for processes and to identify the additional tests needed for stronger claims.

\paragraph{Future work.} These limitations point to work that the dataset and codebook make possible. Applying the final rules to a fresh random sample, with a second coding of the reasons for partial separation and the types of divergence, would give reliability estimates for the final rules, and independent screening would test the eligibility decisions. Recording reference validation for each audit grader would let the census assess grader independence rather than grader type alone. Extending the search to work on similar mechanisms that does not use the RSI label, and repeating the census as the literature grows, would show whether the gaps in audited reruns and same-start comparisons persist.

\section{Conclusion}
\label{sec:conclusion}

This study examined how closely RSI evaluations match the claims made about improvement processes. Product evaluation is widely reported, although fewer studies fully separate it from selection. Most studies claiming accumulating gains audit across rounds, but evaluation of repeatability is less common. Audited reruns, which test whether a loop's gains repeat, appear in only \CRerunsSPct\% of the studies. Similarly, only \CImpClaimSwapMatchedS{} of the \CImpClaimS{} studies claiming a better improver compare earlier and later improvers from identical starting artifacts at a matched budget. Most report related comparisons, but these address different questions. The human-coded sample, the independent improver coding, and the comparison set support this central distinction. The case analysis illustrates why it matters: development gains, audited product gains, and improved procedures can diverge.

The reporting standard derived from these findings connects claims to development signals, audit boundaries, process controls and resource accounting, so that an experiment states what it establishes and which stronger claims require more evidence. Releasing the dataset and codebook will allow the measurements to be repeated as the literature grows and will support a clearer empirical assessment of progress toward better improvers.

\bibliographystyle{unsrtnat}
\bibliography{references}

\appendix
\section{Proofs}
\label{app:proofs}

\paragraph{Proof of Proposition~\ref{prop:null}.}
Write $u_0$ for the common utility and $X_0=e(s_0)$, $X_i=e(c_i)$ for $i=1,\dots,N$. Then $r(s_0)=u_0+X_0$ and $r(c_i)=u_0+X_i$. By induction on $i$, the incumbent after $i$ proposals is the candidate with the largest $X$ among $X_0,\dots,X_i$, because a proposal is accepted exactly when its score exceeds the current incumbent's. Hence proposal $i$ is accepted if and only if $X_i$ is the maximum of $X_0,\dots,X_i$. The variables are exchangeable and ties have probability zero, so this event has probability $1/(i+1)$. Summing over $i=1,\dots,N$ gives $\sum_{j=2}^{N+1}1/j$, which proves (i).

For (ii), $r(S_{\mathrm{final}})-r(s_0)=\max_{0\le i\le N}X_i-X_0\ge 0$. If $X_i\sim\mathcal N(\mu,\sigma^2)$, then $\max_i X_i=\mu+\sigma\max_i Z_i$ with $Z_i$ standard normal, so the expectation is $\sigma m_{N+1}$.

For (iii), $\hat a(S_{\mathrm{final}})-\hat a(s_0)=\alpha(S_{\mathrm{final}})-\alpha(s_0)$ because $u$ is constant. By assumption $\mathbb E[\alpha(S)\mid S]=0$ for every system $S$, so the law of iterated expectations gives $\mathbb E[\alpha(S_{\mathrm{final}})]=\mathbb E[\alpha(s_0)]=0$ and the expected audited gain is zero. The assumption is substantive: it fails, for example, if the audit grader prefers outputs of the kind the development judge selected. The constants in the statement are computed numerically. A Monte Carlo check with 200{,}000 replications agrees with (i) for $N=99$ to the first decimal. $\square$

\paragraph{Proof of Proposition~\ref{prop:coupled}.}
For any system $S$, $\hat a(S)-u(S)=\lambda b(S)+\xi(S)$. Taking the difference between the final and initial systems and then expectations,
\[
\mathbb E[\hat a(S_{\mathrm{final}})-\hat a(s_0)]-\mathbb E[u(S_{\mathrm{final}})-u(s_0)]
=\lambda\,\mathbb E[b(S_{\mathrm{final}})-b(s_0)]+\mathbb E[\xi(S_{\mathrm{final}})]-\mathbb E[\xi(s_0)].
\]
The last two terms vanish because $\mathbb E[\xi(S)\mid S]=0$, which proves the identity for any real $\lambda$.

For the second statement, write $b(S_{\mathrm{final}})-b(s_0)$ as the sum of the changes in $b$ at each accepted change. Each term is at least $\delta$, so the sum is at least $\delta A$, where $A$ is the number of accepted changes. Taking expectations and multiplying by $\lambda\ge 0$ gives the bound. $\square$

\paragraph{Proof of Proposition~\ref{prop:diagonal}.}
Fix an initial distribution $\mu_0$ for the solver state $S_0$. The run is a Markov chain on solver states and gains: $I_t=\phi(S_t)$ and $(S_{t+1},\Delta_t)\sim K(\cdot\mid S_t,\phi(S_t))$. By the Ionescu-Tulcea construction, the joint law of $(S_0,\Delta_0,S_1,\dots,\Delta_{T-1},S_T)$ for every $T$ is determined by $\mu_0$ and the maps $s\mapsto K(\cdot\mid s,\phi(s))$.

Define $\tilde K(\cdot\mid s,I)=K(\cdot\mid s,\phi(s))$ for all $I$. The two kernels agree at every pair the chain visits, so they generate the same law for every $\mu_0$ and every $T$, and hence the same law for any collection of independent runs. Under $\tilde K$ the expected gain $\tilde g(s,I)=g(s,\phi(s))$ does not depend on $I$, so every improver contrast is zero. Under $K$ the contrast $g(s,I')-g(s,\phi(s))$ can take any value. Two models with different contrasts generate the same observables, so the contrast is not identified.

A swap test samples from $K(\cdot\mid s,I')$ with $I'\ne\phi(s)$, which the two models assign different laws whenever their contrasts differ. The frozen arm of an evolving-versus-frozen comparison samples from $K(\cdot\mid s,\phi(S_0))$ only at the solver states $s$ that its own chain visits, so it need not provide the overlap the contrast requires at the states visited by the evolving run. $\square$

\section{Codebook}
\label{app:codebook}

The initial codebook was frozen after the pilot. The v1.1 decision rules below clarify the items revised after the blind coding, the v1.2 rules add three items and the logical constraints between items, and v1.3 records the independent coding of the improver comparisons. Together they govern the reported codes outside the double-coded sample, whose codes were settled by adjudication. The v1.2 and v1.3 rules apply to all studies. Where an earlier rule and a later rule differ, the later rule takes precedence.

The unit is one eligible study, coded from the latest version available on the freeze date. Every non-trivial code carries a locator to a section, table, figure or appendix. The value NR means not reported and is never read as a failed control.

The pilot established a full-text reading procedure: inspect the experimental setup for data splits and hidden tests, search appendices for checkpoint and hyperparameter selection, and examine frozen-improver, fixed-executor, and swap comparisons. It also established that scope must be coded from the authors' claims, and v1.1 specifies which sections govern that judgment. Table~\ref{tab:codebook1} lists the study descriptors and claims, and Table~\ref{tab:codebook2} lists the audit and process items.

\paragraph{Decision rules added after the reliability round (v1.1).}
The blind coding showed systematic differences in how several items were interpreted. We clarified the following rules before recoding the affected items in every study. A report that specifies a loop but runs no experiment on it is excluded as a position paper.

For evaluation separation, A1 is P whenever any headline number includes development, search or selection tasks. A2 is Y only if the held-out tasks were also unused for checkpoint, round, archive-member, hyperparameter and default choices and no ``best observed on test'' value enters the headline. Target-conditioned loops that attempt the evaluation tasks themselves have A2 = N.

For controls, A5 is Y for any control that keeps the loop's budget but removes or randomizes the information that drives selection, including random selection, fixed-proportion selection without feedback, retries without a retained update, and independent sampling from the same start. A6 is Y when candidates are accepted after repeated multi-seed evaluation.

For process evaluation, P1 counts held-out scores at every retained round or checkpoint, including held-out component metrics and periodic checkpoints. P4 is Y for frozen-improver, frozen-evaluator, frozen-curriculum or frozen-operator controls, or earlier-versus-later improvers from a common start, at matched or stated-equal budget, and P for swaps between different improvers or unmatched budgets.

Scope (S1) is judged on the title, abstract and introduction, and a later concession in a limitations section does not remove a scope gap there. B for the reported checkpoint also covers the best archive member chosen on the reported tasks.

\paragraph{Rules added in v1.2.}
A study is eligible if it runs its loop and describes the outcome, quantitatively or qualitatively. A report that specifies a loop and an evaluation protocol but never runs the loop is excluded. Eligibility never depends on the direction of the result.

Four logical rules link the items, with an audit defined as A2 = Y or P. Without an audit, A3 = NA (R1) and P1 = N, because per-round scores are then development scores (R2). Reporting paired development and audit scores (P2 = Y) requires a per-round audit (P1 = Y; R3). Without an audit, P5 = NA because divergence cannot be assessed (R4).

Three items were added. A2r records why an evaluation is only partly separated: R, the evaluation tasks are also used for some selection; H, only part of the headline is held out; U, the paper does not state the separation.

P4d records the design of an improver-related comparison: S, an earlier and a later improver run from identical starting artifacts, or from a declared set of starting states with paired assignment; T, an evolving and a frozen improver each run over the loop; C, another comparison, such as freezing an adaptive evaluator, world model or curriculum rule, or swapping between different models.

Each divergence case is classified as paired, partial, regression or other, as defined in Table~\ref{tab:within}. Evidence tiers are computed from the design items alone, so a study with a negative result keeps the tier of its designs.

\paragraph{Independent coding of improver comparisons (v1.3).}
For every study and anchor work that claims a better improver or has a coded improver-related comparison, P4 and P4d are the codes of an independent blind coder. This coder also recorded the improver, the result of the comparison (P4r: better, mixed, no clear difference or worse for the paper's own process) and a locator.

The strict same-start rule is then applied: equal starting scores do not make starting states equal, and an arm that carries solver artifacts, such as program components or a task agent, into the comparison is not a same-start arm. When a study reports several designs, the first that applies in the order S, T, C is recorded. A comparison with a different improver counts as matched (P4 = Y) when the paper states equal budgets and no other uncontrolled differences. This rule replaces the v1.1 rule that coded every such swap as partial.

\begin{table}[H]
\caption{Study descriptors and claims.}
\label{tab:codebook1}
\centering
\small
\begin{tabular}{p{1.6cm}p{13.2cm}}
\toprule
Item & Values and rule \\
\midrule
Label & Strong: the paper calls its own loop or system RSI or recursively self-improving in the title, abstract or introduction. Weak: RSI appears only as motivation, for example as a step toward it. \\
Self & Y: the changed artifact includes the system that runs the loop (its weights, harness, memory or improver). P: mixed, for example the agent trains a model that it later uses. N: the loop improves a separate target. \\
Target & Any of output, weights, memory or skills, harness or prompt or code, data or curriculum, improver, evaluator, world model. \\
Persist & R: changes are retained and used on later tasks or rounds. T: task-local refinement only. \\
Rounds & Number of rounds, iterations or generations of retained updates, as reported. \\
Domain & V: correctness decided by executable tests, exact answers, simulators, labels or physical measurement. O: open-ended, judged by humans or models. M: both in the headline evaluation. \\
Dev & Signal that ranks or accepts changes: execution or environment, reference labels, humans, a model judge or reward model distinct from the actor, or the actor's own judgment; several allowed. \\
Evmut & Evaluator or criterion changes during the run: fixed; changed by humans; changed by the system with a fixed external reference gating the change; changed by the system without such a reference. \\
Claims & From the abstract, introduction and conclusion: accumulation (gains build or are sustained across rounds); improver (the improvement process itself improves); acceleration. \\
Result & Positive, negative (the study reports that its loop did not deliver the intended improvement) or mixed. \\
Quant & Y: the change is evaluated quantitatively. N: qualitative demonstration only (tier E0). \\
\bottomrule
\end{tabular}
\end{table}

{\small
\begin{longtable}{p{0.7cm}p{3.3cm}p{1.3cm}p{9.3cm}}
\caption{Audit and process items.}
\label{tab:codebook2}\\
\toprule
Item & Name & Values & Rule \\
\midrule
\endfirsthead
\multicolumn{4}{l}{\small Table~\thetable{} (continued)}\\
\toprule
Item & Name & Values & Rule \\
\midrule
\endhead
\bottomrule
\endfoot
A1 & Headline on development signal & Y, P, N & Y if the headline number is measured with the tasks or signal that selected or accepted the changes; P if it mixes both, or uses the same tasks with hidden tests. \\
A2 & Product evaluation separation & Y, P, N, NR & Y requires tasks never used to select or accept changes and not used to pick checkpoints, rounds or hyperparameters. P if held out but also used for such choices, if overlap is unclear, or if only part of the headline is held out; the reason is recorded in A2r. \\
A3 & Audit grader & E, L, H, J, JS, M, NA & Execution or simulator; labels; human raters; model judge of another family; model judge of the same family or the system's own evaluator; mixed; no audit (A2 = N or NR). Type only, not validity. \\
A4 & Budget-matched comparator & Y, P, N & Y: a baseline without the recursive update at an explicitly matched budget, such as best-of-$N$, a fixed improver or more search. P: budgets reported or roughly comparable but not matched. \\
A5 & Null control & Y, N & Shuffled feedback, inert proposals, random acceptance or a no-feedback loop. \\
A6 & Noise-aware promotion & Y, N & Acceptance uses significance tests, confidence bounds, repeated evaluation or a margin justified by noise. \\
U & Uncertainty reported & Y, N & Any variance, interval or bootstrap for the headline result. \\
Ckpt & Reported checkpoint & F, V, B, NR & Final or pre-declared; selected on validation; best on the reported benchmark. \\
P1 & Per-round audit & Y, N & Audit-grade score at each retained round or at scheduled checkpoints, including held-out metrics of an updated component; requires an audit. \\
P2 & Development and audit per round & Y, N & Both reported for the same rounds. \\
P3 & Loop reruns & count, NR & Independent reruns of the full loop; evaluation seeds of one evolved system do not count. \\
P3a & Audited reruns & Y, N & At least two reruns report audit-level outcomes. \\
P4 & Improver-related comparison & Y, P, N & Y: an earlier/later improver comparison from a common start, a frozen-improver or other frozen-component control, or a comparison with a different improver, at a matched or stated-equal budget with no other uncontrolled differences; a properly designed inconclusive comparison counts. P: partial, for example an unmatched budget or a comparison confounded with other differences. The design is recorded in P4d and, for independently coded studies, the result in P4r. Coded for every study. \\
P5 & Reported divergence & Y, P, N, NA & The study reports development gains that do not carry to the audit, or that shrink or regress while the development signal rises. N: both are reported and agree. NA: no audit is paired with development scores. \\
S1 & Scope gap & Y, P, N & The claimed domain or level exceeds what is tested. \\
\end{longtable}
}

\section{Search Log}
\label{app:search}

We queried the arXiv API on \FrameFreezeDate{} with each phrase in the title field and in the abstract field, restricted to submissions up to that date, and took the union of the results. The arXiv search normalizes word forms, so the inflected variants return the same records, but we keep them in the log for transparency. Table~\ref{tab:search} lists the queries.

Of the \FrameTotal{} records, \FrameTwentySix{} were posted in 2026 and \FramePreTwentySix{} earlier. Posting accelerated sharply during 2026: \FrameFirstHalfTwentySix{} records appeared in the first half of the year, followed by \FrameMonthJul{} in July, \FrameMonthAug{} in August, \FrameMonthSep{} in September and \FrameMonthOct{} in the first seventeen days of October. These counts include surveys, position papers and other records that the eligibility rules exclude.

\begin{table}[H]
\caption{Search queries and hits on \FrameFreezeDate.}
\label{tab:search}
\centering
\small
\begin{tabular}{llr}
\toprule
Field & Phrase & Hits \\
\midrule
title & recursive self-improvement & 85 \\
abstract & recursive self-improvement & 161 \\
title & recursive self improvement & 85 \\
abstract & recursive self improvement & 161 \\
title & recursively self-improving & 85 \\
abstract & recursively self-improving & 161 \\
title & recursive self-improving & 85 \\
abstract & recursive self-improving & 161 \\
title & recursively self-improve & 85 \\
abstract & recursively self-improve & 161 \\
title & recursively improve itself & 0 \\
abstract & recursively improve itself & 0 \\
\midrule
Union of all queries & & 173 \\
\bottomrule
\end{tabular}

\end{table}

\section{Per-Study Codes}
\label{app:codes}

This appendix lists the main codes for every eligible study in the frame (Table~\ref{tab:perstudy}). These are adjudicated codes for the double-coded sample, independent codes for the improver comparisons, and v1.1 codes otherwise, with the v1.2 rules and the strict same-start rule applied throughout. The remaining tables support three parts of the analysis. Tables~\ref{tab:matrix} and~\ref{tab:within} report evaluation items by domain and the documented divergence cases. Tables~\ref{tab:agreement} and~\ref{tab:improveraudit} report agreement for the double-coded sample and the independent improver coding. Tables~\ref{tab:anchor}, \ref{tab:anchorcodes}, and~\ref{tab:wuconc} present the anchor comparison, the individual anchor codes, and their correspondence with the protocol descriptions of \citet{wu2026rsisurvey}. The full codes, with locators and notes for every item, will be released with the paper.

{\small
\setlength{\tabcolsep}{4pt}
\begin{longtable}{l>{\raggedright\arraybackslash}p{5.9cm}cccccccccc}
\caption{Codes for every eligible study (S = strong label, W = weak label; domain V = verifiable, O = open-ended, M = mixed). $^{\ast}$Double-coded and adjudicated. Design of the improver comparison: S = same start, earlier versus later improver; T = evolving versus frozen improver over a run; C = other component frozen, or different improvers. Item definitions are in Appendix~\ref{app:codebook}.}\label{tab:perstudy}\\
\toprule
arXiv & Title & Lab. & Dom. & A2 & P1 & P3a & A4 & A5 & P4 & Design & Tier \\
\midrule
\endfirsthead
\toprule
arXiv & Title & Lab. & Dom. & A2 & P1 & P3a & A4 & A5 & P4 & Design & Tier \\
\midrule
\endhead
\bottomrule
\endfoot
1312.6764 & Bounded Recursive Self-Improvement & S & M & NR & N & N & N & N & N & -- & E0 \\
2202.05780 & A Modern Self-Referential Weight Matrix Tha... & W & V & Y & N & Y & P & N & N & -- & E2 \\
2310.02304 & Self-Taught Optimizer (STOP): Recursively S... & S & V & Y & Y & Y & P & N & Y & S & E4 \\
2410.04444$^{\ast}$ & Gödel Agent: A Self-Referential Agent Frame... & S & V & Y & N & N & Y & Y & P & C & E2 \\
2410.12375 & PRefLexOR: Preference-based Recursive Langu... & W & O & NR & N & N & N & N & N & -- & E1 \\
2410.15639 & Can Large Language Models Invent Algorithms... & S & V & P & Y & N & P & N & P & S & E3 \\
2502.09963 & Generating on Generated: An Approach Toward... & S & O & Y & Y & N & P & Y & N & -- & E3 \\
2602.15659 & Can Recommender Systems Teach Themselves? A... & S & V & Y & Y & N & P & N & N & -- & E3 \\
2603.02203 & Tool Verification for Test-Time Reinforceme... & W & V & P & Y & Y & P & N & N & -- & E3 \\
2603.06333 & SAHOO: Safeguarded Alignment for High-Order... & S & M & P & Y & N & N & N & N & -- & E2 \\
2603.23129 & Polaris: A Gödel Agent Framework for Small... & S & V & Y & N & Y & P & N & P & C & E3 \\
2605.15871$^{\ast}$ & Agentic Discovery of Neural Architectures:... & W & V & P & N & Y & P & N & N & -- & E3 \\
2606.20657 & A-Evolve-Training: Autonomous Post-Training... & W & V & P & Y & N & N & N & N & -- & E3 \\
2606.26294 & The Red Queen Gödel Machine: Co-Evolving Ag... & S & M & Y & Y & N & Y & N & Y & C & E4 \\
2607.05297 & MetaSkill-Evolve: Recursive Self-Improvemen... & S & V & Y & N & N & Y & N & P & T & E2 \\
2607.05382$^{\ast}$ & Search Beyond What Can Be Taught: Evolving... & W & O & Y & Y & N & P & N & N & -- & E3 \\
2607.08252 & A Multi-Timescale Recursive Self-Improvemen... & S & O & N & N & N & P & N & N & -- & E1 \\
2607.12625 & KnowAct-GUIClaw: Know Deeply, Act Perfectly... & W & M & P & N & N & P & N & N & -- & E2 \\
2607.15314 & Cura 1T: Healthcare Foundation Model via Re... & S & M & N & N & N & N & N & N & -- & E1 \\
2607.21461 & AREX: Towards a Recursively Self-Improving... & S & V & Y & N & N & Y & N & N & -- & E2 \\
2607.28251$^{\ast}$ & Policy-Level Recursive Self-Improvement for... & S & V & P & Y & N & P & Y & N & -- & E3 \\
2607.28568 & Frontis-MA1: Training an AI4AI Model toward... & W & V & Y & N & N & Y & N & Y & S & E2 \\
2608.07645$^{\ast}$ & Mendel Gödel Machine: Recursive Self-Improv... & S & V & P & N & N & P & N & N & -- & E2 \\
2608.09819$^{\ast}$ & Macaron-V1: Towards Open Continual Learning... & S & M & P & N & N & P & Y & N & -- & E2 \\
2608.10157 & SBCO: Self-Supervised, Verifier-Grounded Ha... & W & V & P & N & N & P & N & N & -- & E2 \\
2608.12841 & AQuA: Recursively Self-Improving Quantitati... & S & V & Y & N & N & N & N & N & -- & E2 \\
2608.13560 & AutoDesign: Meta-Harness Optimization for L... & W & O & P & N & N & P & N & N & -- & E2 \\
2608.13951 & HELIX: Model-Harness Co-evolution for Recur... & S & V & P & N & N & P & N & N & -- & E2 \\
2608.14621 & AutoMem: A Text-Gradient Recursive Self-Imp... & S & V & Y & N & N & P & Y & N & -- & E2 \\
2608.16211 & BaT: Towards Self-Evolving Medical Research... & S & M & P & Y & N & P & N & N & -- & E3 \\
2608.24231 & RecurSE: Bounded Recursive Self-Evaluation... & S & V & Y & Y & N & Y & N & P & C & E3 \\
2608.24735 & Metan: Recursive Self-Improvement through E... & S & V & P & N & Y & Y & N & Y & C & E4 \\
2608.24876$^{\ast}$ & Recursive Experiential-Working Memory Evolu... & S & V & P & Y & Y & Y & Y & P & C & E3 \\
2608.26334 & ProofEvolve: Neuro-Symbolic Evolution for F... & W & V & Y & N & Y & Y & Y & N & -- & E3 \\
2608.26418 & Redwood: A Frontier AI Accelerator Designed... & S & V & NR & N & N & N & N & N & -- & E0 \\
2608.27266$^{\ast}$ & Naive Prompt Optimization: Rethinking the N... & W & V & P & N & N & P & N & Y & C & E2 \\
2608.29966 & DataFoundry: Evolving Data Preparators via... & S & M & Y & Y & N & P & N & N & -- & E3 \\
2609.06396 & MetaRSI / RSI2: A Meta-Recursive Self-Impro... & S & V & Y & N & Y & Y & N & Y & S & E4 \\
2609.08183 & NeoHorse-1: Towards Recursive Self-Improvem... & W & M & P & N & N & P & N & N & -- & E2 \\
2609.10702 & Data-Efficient Language Modeling: From Fron... & S & V & P & N & N & P & N & N & -- & E2 \\
2609.14857$^{\ast}$ & ModularRSI: Modular and Generalizable Recur... & S & V & Y & Y & N & Y & N & N & -- & E3 \\
2609.14858 & Dream-RSI: Recursive Self-Improvement throu... & S & V & P & N & N & Y & N & Y & T & E2 \\
2609.15364$^{\ast}$ & RSIAgent: Autonomous Exploration for Recurs... & S & M & N & N & N & P & N & N & -- & E1 \\
2609.17523 & ScienceBuddy: Recursive-in-Recursive Self-I... & S & O & Y & N & N & P & N & N & -- & E2 \\
2609.19526 & Self Improvement via Fast Tree-search & S & V & P & N & N & Y & N & N & -- & E2 \\
2609.20519 & SoL-Pi: Recursively Scaling Auto-Research L... & W & V & Y & N & N & P & N & N & -- & E2 \\
2609.24838 & MedRSI: Recursive Self-Improvement for Medi... & S & V & Y & Y & Y & P & N & N & -- & E3 \\
2609.24972 & RRSI: Regularized Recursive Self-Improvemen... & S & M & Y & N & N & P & N & N & -- & E2 \\
2609.26457 & Recursive self-improvement of AI research a... & S & V & Y & N & N & Y & N & Y & S & E2 \\
2609.26863 & Towards Strategy-Level RSI for Skill-Augmen... & S & V & Y & Y & N & P & Y & N & -- & E3 \\
2609.27389$^{\ast}$ & EvoAudio: Recursive Self-Improvement for Au... & S & V & Y & Y & N & Y & N & Y & C & E4 \\
2609.29626 & iCoder-27B: Recursive AI-Led Development of... & S & V & P & N & N & P & Y & N & -- & E2 \\
2609.30652 & Recursive Self-Improvement via On-Policy Di... & S & V & P & Y & N & Y & N & Y & T & E4 \\
2609.31760 & What Stops Recursive Self-Improvement in Ro... & S & V & P & Y & N & N & N & N & -- & E3 \\
2609.32984 & ReVision3D: Attribution-Guided Recursive Se... & S & V & Y & N & N & P & N & N & -- & E2 \\
2609.33386$^{\ast}$ & Sivia: Learning to Draw, Learning to Learn & S & O & Y & Y & N & Y & N & Y & S & E4 \\
2609.33398$^{\ast}$ & COEVO: Co-Evolving Context and Parameters f... & S & V & Y & N & N & P & N & N & -- & E2 \\
2609.33867 & R2 Flow: Recursive Self-Improvement via Rec... & S & M & Y & Y & Y & P & Y & N & -- & E3 \\
2609.34712 & RSI-Router: Evolving Subtask-Level LLM Rout... & S & V & Y & N & N & P & N & N & -- & E2 \\
2609.35110 & Sol-H3: Recursive Self-Improvement for Mini... & S & V & P & N & N & P & N & N & -- & E2 \\
2609.35328$^{\ast}$ & Hyper Algorithm Design Agent: Evolving Lear... & S & V & P & N & N & Y & N & Y & C & E2 \\
2609.35561$^{\ast}$ & RSI-Master: Structuring Experiments to Guid... & S & M & N & N & N & P & N & N & -- & E1 \\
2609.36235 & MERID: Multimodal Exploration via Recursive... & S & V & Y & Y & Y & P & N & N & -- & E3 \\
2609.36549 & MedForge-RSI: Medical Deepfake Detection vi... & S & V & Y & Y & Y & P & N & N & -- & E3 \\
2609.36675$^{\ast}$ & Gödel Forest: Balancing Search Depth and Br... & S & V & N & N & N & P & N & N & -- & E1 \\
2609.36695$^{\ast}$ & Know Thyself, Teach Thyself: Internal Infor... & W & V & P & Y & N & P & Y & N & -- & E3 \\
2609.37834$^{\ast}$ & Mixture of Self-Improving Branches For Agen... & S & V & P & Y & N & Y & N & Y & T & E4 \\
2609.37950 & Video-RSI: Recursive Self-Improvement of Vi... & S & V & Y & Y & N & P & N & N & -- & E3 \\
2609.38372 & Self-Evolving Harness on Multiple Tasks wit... & S & M & Y & N & N & P & N & N & -- & E2 \\
2609.38662$^{\ast}$ & CollabFlow: Recursive Self-Improvement of A... & S & V & Y & N & N & Y & Y & Y & S & E2 \\
2609.38721$^{\ast}$ & UniEvo-VL: An On-policy Self-Distillation T... & W & M & P & Y & N & P & N & P & C & E3 \\
2609.38905$^{\ast}$ & EmbodiRSI: Recursive Self-Improvement for D... & S & V & Y & Y & N & Y & Y & N & -- & E3 \\
2609.39045 & RSIGame: Autonomous Agentic Game Developmen... & S & O & Y & Y & N & Y & N & N & -- & E3 \\
2609.39306 & ReSAIL: Mitigating Collapse in Iterative Ag... & W & V & Y & Y & N & Y & N & N & -- & E3 \\
2609.40330$^{\ast}$ & Turbo Harness: Instance-Adaptive Harness Op... & W & V & Y & N & N & P & N & Y & S & E2 \\
2610.03002 & Recursive Self-Improvement in Unified Multi... & S & V & Y & Y & N & Y & Y & Y & C & E4 \\
2610.04302 & From Valid to Useful: Post-Verification Acq... & S & V & Y & Y & Y & Y & N & N & -- & E3 \\
2610.04927 & Assembling Insights for Agentic Machine Lea... & W & V & Y & N & Y & P & N & N & -- & E3 \\
2610.05151 & Recursive Self-Improvement of Visuomotor Po... & S & V & P & N & N & Y & N & N & -- & E2 \\
2610.05701 & Second-Order Problem Solving for Recursive... & S & V & Y & Y & Y & Y & N & N & -- & E3 \\
2610.07105 & Beyond Successor Accuracy: State Retention... & S & V & Y & N & Y & Y & Y & N & -- & E3 \\
2610.07237 & SPECTRUM: Proximal Spectral Modulation for... & W & V & Y & Y & N & P & N & N & -- & E3 \\
2610.07969 & EmbodiedSmith: Scaling Embodied Data throug... & S & M & Y & N & N & P & Y & N & -- & E2 \\
2610.08640$^{\ast}$ & RIWANav: Recursive World-Action Models with... & S & V & Y & Y & Y & P & N & P & C & E3 \\
2610.08995 & PhysEvo: Astra Can Act, Let It & S & V & Y & N & N & P & N & N & -- & E2 \\
2610.11794 & Memento 3: Model-Based Recursive Self-Impro... & S & V & P & N & N & P & N & N & -- & E2 \\
2610.12176 & Recursive Self-Improvement through Multi-Ag... & S & O & Y & Y & N & P & N & Y & S & E4 \\
2610.12233 & ReSI: Recursive Safety Improvement toward R... & S & M & P & N & N & P & N & N & -- & E2 \\
2610.12369 & Embodied Turing Machines: Stateful Code for... & S & V & Y & N & N & P & N & N & -- & E2 \\
2610.13098$^{\ast}$ & VETR: Execution-Grounded Self-Improving Age... & S & V & N & N & N & P & N & N & -- & E1 \\
2610.13376 & EndoRSI: Physical-Evidence-Driven Recursive... & S & V & Y & Y & Y & P & N & N & -- & E3 \\
2610.13912 & RSI-Anything: Unlocking Recursive Self-Impr... & S & M & Y & Y & Y & Y & Y & Y & C & E4 \\
2610.14163 & AutoEnvScaling: Automating the Data Flywhee... & W & V & Y & N & N & Y & Y & N & -- & E2 \\
2610.14224 & Harness\^{}2: Recursive Agent Harnessing for a... & S & M & Y & N & Y & Y & N & N & -- & E3 \\
2610.14843 & Pre-Registered Genesis Accounting for Recur... & S & V & Y & N & N & N & Y & N & -- & E2 \\
2610.15708$^{\ast}$ & SpecForge: Recursive Self-Improvement for C... & S & V & P & Y & N & P & N & N & -- & E3 \\
2610.15772 & RSIPO: Recursive Self-Improving Policy Opti... & S & V & Y & N & N & P & N & N & -- & E2 \\
2610.15782 & Data-Centric Recursive Self-Improvement for... & S & V & P & N & N & P & N & N & -- & E2 \\
2610.16078$^{\ast}$ & Heterogeneous Agentic Coevolution & S & V & Y & N & Y & P & N & P & T & E3 \\
2610.16491 & Drafting Agents, Not Just Tokens: Speculati... & S & V & P & Y & Y & Y & N & P & S & E3 \\
2610.17114$^{\ast}$ & D-World: An Explicit World-Language-Action... & S & V & N & N & N & N & N & N & -- & E1 \\
2610.17718$^{\ast}$ & ECG-RSI: Recursive Self-Improvement through... & S & V & Y & Y & N & Y & Y & Y & T & E4 \\
2610.18083 & AReaL-RSI: Adaptive Scaling and Coordinatio... & S & M & P & N & N & N & N & N & -- & E2 \\
2610.18097 & APT: Feedback-Driven Agentic Post-Training... & W & O & Y & N & N & P & Y & N & -- & E2 \\
2610.18845 & The Last Remote Sensing Model Built by Huma... & S & V & P & Y & N & P & N & N & -- & E3 \\
2610.19424 & VideoHarness-SRSI: Structured Recursive Sel... & S & V & Y & N & Y & Y & N & N & -- & E3 \\
2610.20431 & SwarmRSI: Recursive Self-Improvement throug... & S & M & Y & Y & Y & Y & N & Y & C & E4 \\
2610.20474 & AdaRSI: On-Demand Recursive Self-Improvemen... & S & V & Y & N & N & Y & N & N & -- & E2 \\
2610.20665 & The World Won't Wait: Evaluating and Teachi... & S & V & Y & N & N & N & N & N & -- & E2 \\
2610.21040 & Recursive Reward Improvement via Adversaria... & S & O & Y & N & N & Y & N & Y & T & E2 \\
2610.21055$^{\ast}$ & Athanor: An Agent-Native System for Autonom... & W & V & P & N & N & P & N & N & -- & E2 \\
2610.21791 & RSIFlow: Adaptive Self-Improvement Workflow... & S & V & Y & Y & N & N & N & P & S & E3 \\
2610.22192 & Terminal-RSI: Recursive Self-Improving Term... & S & V & P & Y & N & N & N & N & -- & E3 \\
2610.22344 & MLExpert: Towards a Generalist Agentic Mode... & W & V & Y & N & N & P & N & N & -- & E2 \\
\end{longtable}

}

\begin{table}[H]
\caption{Evidence items by domain. Counts are studies, with the share of the column in parentheses. Verifiable domains decide correctness with tests, exact answers, simulators or labels; open or mixed domains include outputs judged by humans or models. The last column repeats the first for all eligible studies, including the weak-label ones.}
\label{tab:matrix}
\centering
\small
\resizebox{\textwidth}{!}{\begin{tabular}{p{6.2cm}cccc}
\toprule
Item & Strong label ($n=90$) & Verifiable ($n=65$) & Open or mixed ($n=25$) & All eligible ($n=114$) \\
\midrule
\multicolumn{5}{l}{Product audit} \\
Selection-independent evaluation (A2 = Y) & 54 (60\%) & 39 (60\%) & 15 (60\%) & 66 (58\%) \\
Partly separated evaluation (A2 = P) & 27 (30\%) & 22 (34\%) & 5 (20\%) & 38 (33\%) \\
\quad evaluation also used for selection & 14 (16\%) & 11 (17\%) & 3 (12\%) & 19 (17\%) \\
\quad only part of the headline held out & 5 (6\%) & 5 (8\%) & 0 (0\%) & 8 (7\%) \\
\quad separation not stated & 8 (9\%) & 6 (9\%) & 2 (8\%) & 11 (10\%) \\
No audit beyond development tasks (A2 = N or NR) & 9 (10\%) & 4 (6\%) & 5 (20\%) & 10 (9\%) \\
Headline measured on development signal (A1 = Y) & 8 (9\%) & 5 (8\%) & 3 (12\%) & 9 (8\%) \\
Best checkpoint chosen on reported benchmark & 11 (12\%) & 7 (11\%) & 4 (16\%) & 15 (13\%) \\
Any uncertainty reported for headline (U = Y) & 46 (51\%) & 35 (54\%) & 11 (44\%) & 60 (53\%) \\
\midrule
\multicolumn{5}{l}{Controls for the recursion} \\
Budget-matched comparator (A4 = Y) & 33 (37\%) & 26 (40\%) & 7 (28\%) & 37 (32\%) \\
Null or no-feedback control (A5 = Y) & 17 (19\%) & 12 (18\%) & 5 (20\%) & 21 (18\%) \\
Noise-aware promotion rule (A6 = Y) & 17 (19\%) & 12 (18\%) & 5 (20\%) & 21 (18\%) \\
\midrule
\multicolumn{5}{l}{Process evidence} \\
Audits across rounds or checkpoints (P1 = Y) & 39 (43\%) & 28 (43\%) & 11 (44\%) & 46 (40\%) \\
Development and audit at each round (P2 = Y) & 10 (11\%) & 6 (9\%) & 4 (16\%) & 11 (10\%) \\
Two or more audited reruns of the loop (P3a = Y) & 20 (22\%) & 16 (25\%) & 4 (16\%) & 25 (22\%) \\
Improver comparison at matched budget (P4 = Y) & 18 (20\%) & 12 (18\%) & 6 (24\%) & 21 (18\%) \\
\quad same start, earlier versus later improver & 6 (7\%) & 4 (6\%) & 2 (8\%) & 8 (7\%) \\
\quad evolving versus frozen improver over a run & 5 (6\%) & 4 (6\%) & 1 (4\%) & 5 (4\%) \\
\quad other component frozen, or different improvers & 7 (8\%) & 4 (6\%) & 3 (12\%) & 8 (7\%) \\
Reports development--audit divergence (P5 = Y or P) & 22 (24\%) & 16 (25\%) & 6 (24\%) & 24 (21\%) \\
\midrule
\multicolumn{5}{l}{Claims} \\
Claims the improver itself improves & 26 (29\%) & 20 (31\%) & 6 (24\%) & 27 (24\%) \\
\quad and compares earlier and later improvers (P4 = Y, same start) & 4 (4\%) & 2 (3\%) & 2 (8\%) & 4 (4\%) \\
Claims gains accumulate across rounds & 45 (50\%) & 31 (48\%) & 14 (56\%) & 46 (40\%) \\
\quad and audits across rounds or checkpoints (P1 = Y) & 27 (30\%) & 17 (26\%) & 10 (40\%) & 28 (25\%) \\
Claimed scope exceeds tested scope (S1 = Y or P) & 27 (30\%) & 18 (28\%) & 9 (36\%) & 32 (28\%) \\
\bottomrule
\end{tabular}
}
\end{table}

\begin{table}[p]
\caption{Every eligible study that reports, in its own experiments, development gains that the audit did not confirm in full (P5 = Y or P). Type: paired, development signal and audit disagree about the same systems; partial, smaller or shifted-domain transfer; regression, gains reverse across rounds or fail to replicate; other, the audit was compromised. The subject is the proposed method unless marked as a variant (ablation or unpromoted candidate) or a baseline. $^{\dagger}$Weak-label study. Locators refer to the version coded on \FrameFreezeDate.}
\label{tab:within}
\centering
\footnotesize
\setlength{\tabcolsep}{3pt}
\begin{tabular}{>{\raggedright\arraybackslash}p{2.2cm}>{\raggedright\arraybackslash}p{1.6cm}>{\raggedright\arraybackslash}p{3.5cm}>{\raggedright\arraybackslash}p{6.0cm}>{\raggedright\arraybackslash}p{2.1cm}}
\toprule
Study & Type & Development signal & What the study's own audit or control showed & Locator \\
\midrule
\citet{wang2026robotics} & Paired & step-level gates with privileged state; paired trials & 123 rounds; target task never succeeded; the reach evaluator was gamed & Abstract; Fig. 1b; Fig. 3 \\
\citet{wang2026ecgrsi} & Paired & feedback-set error (39.63\% to 21.33\%) & external gains 13.07, 3.01, 0.38 points; three benchmarks regress in round 3 & Sec. 5.7; Tab. 18 \\
\citet{shi2026aevolve}$^{\dagger}$ & Paired & public leaderboard used inside the loop & development metric reached record highs without moving the external target & Abstract; Sec. 3; Fig. 3 \\
\citet{lee2026mass} & Paired & self-evaluation (same model) & judge-graded gains; executable benchmarks flat (37.5 to 35.2, 62.6 to 61.1) & Sec. 4; Fig. 2 \\
\citet{pepe2026aira}$^{\dagger}$ & Paired & validation fitness during search & best validation and best test fitness separate over search steps; designs ranked on test accuracy & App B; Fig. 1c; Fig. 8 \\
\citet{fu2026sift} & Paired & search-time accuracy on a 50-task subset & search-time accuracy does not reliably identify the best agent under full evaluation & Sec. 4; Tab. 5 \\
\citet{xia2026rrsi} & Paired (variant) & evolution-split score & unregularized arm: highest evolution score (92.8), out-of-distribution average 40.3, near unevolved & Fig. 1a; Tab. 2 \\
\citet{dong2026branches} & Paired (baseline) & branch development subsets & development rankings only partly carry to the test split: test-based selection would raise the Meta-Harness arm from 46.0\% to 49.0\% and from 29.0\% to 32.5\% & Sec. 4.6; Tab. 4 \\
\citet{zhang2026r2flow} & Paired (variant) & training reward from reward-derived labels & most admitted edits raise training reward while lowering held-out verified score & Sec. 5; Fig. 5c--d \\
\citet{lin2026videoharness} & Paired (baseline) & validation accuracy & validation keeps rising while held-out accuracy does not; method cuts the gap by 37.9\% & Sec. 3; Sec. 5.3 \\
\citet{chen2026coevo} & Paired (variant) & on-policy training performance & variant with stronger training performance reaches lower held-out scores & Sec. 5.3; App. C.2 \\
\citet{liu2026recurse} & Partial & same-style validation accuracy & ``validation deception'': same-style accuracy rises, cross-style transfer degrades & Sec. 5; Tab. 2--3 \\
\citet{luo2026videorsi} & Partial & selection-question accuracy & selection gains need not transfer at every step on a held-out benchmark & Sec. 4.4; Fig. 3 \\
\citet{liu2026mendel} & Partial & accuracy on the 60 Polyglot evolution tasks (50.8\% to 93.2\%) & gains shrink on held-out SWE-bench Pro and Multilingual (+10.0 and +13.3 points); the HGM-evolved baseline loses 3.4 points on SWE-bench Pro & Sec. 5.2; Tab. 2 \\
\citet{wang2026evoaudio} & Partial & accuracy on a fixed held-out promotion set & ``a candidate that gains on the held out set does not always gain on the benchmarks'' & Sec. 3.3; Fig. 2 \\
\citet{yang2026terminalrsi} & Partial & training reward on verified trajectories & continued training keeps Terminal-Bench 2.1 but loses 2.4 to 3.0 points on Terminal-Bench 3.0, 4.0 and Hard & Sec. 3.3 \\
\citet{chen2026medforge} & Regression & two-stage acceptance on development folds & best single-run result often fails to replicate across seeds & Sec. 6.2; App. G \\
\citet{yu2026rewm} & Regression & validation gate & one run plateaus and another reverses on held-out tasks & Sec. 3.4.2; Sec. 3.5 \\
\citet{zhang2025generating} & Regression & human-preference scorers & gains slow after round 3; one scorer declines in some configurations & Sec. 4.3; Tab. 1 \\
\citet{yang2026datafoundry} & Regression & pilot-set data-quality assessment & best at round 2; Finance falls from 66.1 to 58.3 after one more round & Sec. 5.1; Fig. 3 \\
\citet{xu2026selfharness} & Regression & scores on evolution batches & iteration 5 scores below the seed on two held-out benchmarks & Sec. 5.6 \\
\citet{hasan2026darsir} & Regression & fidelity-verified acquisition & performance peaks at round 4 and declines at round 5 & Fig. 3 \\
\citet{lee2026revision3d} & Regression (variant) & short-trial gain above seed range & short-trial gains not retained at 400 steps (not promoted) & Sec. 4; App. G.2 \\
\citet{srikanth2026aide} & Other & private grade on selection tasks & evolved agent patched a held-out scoring script; swap test inconclusive & Sec. 3.5--3.6; Fig. 7 \\
\bottomrule
\end{tabular}

\end{table}

\begin{table}[H]
\caption{Agreement between the blind first-pass codes and the independent human second coding on \RelN{} randomly sampled eligible studies, before adjudication. Left: items as coded. Right: the binary items that enter the analysis. Cohen's $\kappa$ is undefined when expected agreement equals one.}
\label{tab:agreement}
\centering
\small
\resizebox{\textwidth}{!}{\begin{tabular}{p{6.2cm}ccp{0.4cm}p{5.2cm}cc}
\toprule
Item as coded & Agreement & $\kappa$ & & Binary item used in the analysis & Agreement & $\kappa$ \\
\midrule
Strong or weak label (label) & 97\% & 0.91 & & Fully or partly held-out evaluation (A2 = Y or P) & 87\% & 0.44 \\
Domain (domain) & 97\% & 0.90 & & Selection-independent evaluation (A2 = Y) & 67\% & 0.36 \\
Changed artifact includes the loop's system (self) & 90\% & 0.82 & & Per-round audit (P1 = Y) & 80\% & 0.57 \\
Changes persist (persist) & 97\% & 0.65 & & Audited reruns (P3a = Y) & 97\% & 0.87 \\
Accumulation claim (cl acc) & 83\% & 0.63 & & Budget-matched comparator (A4 = Y) & 67\% & 0.25 \\
Improver claim (cl imp) & 93\% & 0.81 & & Null control (A5 = Y) & 80\% & 0.33 \\
Uncertainty reported (U) & 93\% & 0.86 & & Noise-aware promotion (A6 = Y) & 80\% & 0.14 \\
Headline on development signal (A1) & 60\% & 0.30 & & Matched improver comparison (P4 = Y) & 83\% & 0.35 \\
Held-out product audit (A2) & 50\% & 0.18 & & Tier E4 & 83\% & 0.19 \\
Audit grader (A3) & 63\% & 0.53 & &  & &  \\
Budget-matched comparator (A4) & 60\% & 0.24 & &  & &  \\
Null control (A5) & 80\% & 0.33 & &  & &  \\
Noise-aware promotion (A6) & 80\% & 0.14 & &  & &  \\
Per-round audit (P1) & 80\% & 0.57 & &  & &  \\
Audited reruns (P3a) & 97\% & 0.87 & &  & &  \\
Improver test (P4) & 63\% & 0.22 & &  & &  \\
Reported divergence (P5) & 70\% & 0.26 & &  & &  \\
Scope gap (S1) & 53\% & 0.12 & &  & &  \\
Reported checkpoint (ckpt) & 57\% & 0.41 & &  & &  \\
Evaluator changes (evmut) & 73\% & 0.20 & &  & &  \\
\bottomrule
\end{tabular}
}
\end{table}

\begin{table}[H]
\caption{Independent coding of improver-related comparisons for the \AuditFourN{} studies and anchor works that claim a better improver or had been coded with such a comparison, before resolution. Rows are our codes (v1.2), columns the independent codes. The final codes adopt the independent codes and apply the strict same-start rule, which moves two same-start comparisons to other designs.}
\label{tab:improveraudit}
\centering
\small
\begin{tabular}{lccc}
\toprule
\multicolumn{4}{c}{Comparison (P4)} \\
Ours $\downarrow$ / independent $\rightarrow$ & Y & P & N \\
\midrule
Y & 20 & 3 & 0 \\
P & 7 & 7 & 0 \\
N & 0 & 2 & 4 \\
\bottomrule
\end{tabular}\hspace{1.5em}
\begin{tabular}{lcccc}
\toprule
\multicolumn{5}{c}{Design (P4d)} \\
Ours $\downarrow$ / independent $\rightarrow$ & Same start & Evolving vs frozen & Other & None \\
\midrule
Same start & 9 & 0 & 0 & 0 \\
Evolving vs frozen & 2 & 10 & 2 & 0 \\
Other & 2 & 0 & 12 & 0 \\
None & 0 & 0 & 2 & 4 \\
\bottomrule
\end{tabular}

\end{table}

\begin{table}[H]
\caption{Strong-label studies in the arXiv RSI frame against the anchor works, the twenty studies that \citet{wu2026rsisurvey} select as representative of update mechanisms. Per-work codes are in Table~\ref{tab:anchorcodes}.}
\label{tab:anchor}
\centering
\small
\begin{tabular}{p{7.4cm}cc}
\toprule
Item & Strong-label studies ($n=90$) & Anchor works ($n=20$) \\
\midrule
Selection-independent evaluation (A2 = Y) & 54 (60\%) & 9 (45\%) \\
Fully or partly held-out evaluation (A2 = Y or P) & 81 (90\%) & 19 (95\%) \\
Headline measured on development signal (A1 = Y) & 8 (9\%) & 4 (20\%) \\
Audits across rounds or checkpoints (P1 = Y) & 39 (43\%) & 6 (30\%) \\
Two or more audited reruns (P3a = Y) & 20 (22\%) & 5 (25\%) \\
Budget-matched comparator (A4 = Y) & 33 (37\%) & 4 (20\%) \\
Null or no-feedback control (A5 = Y) & 17 (19\%) & 4 (20\%) \\
Noise-aware promotion rule (A6 = Y) & 17 (19\%) & 0 (0\%) \\
Any uncertainty for headline (U = Y) & 46 (51\%) & 10 (50\%) \\
Reaches tier E4 & 13 (14\%) & 5 (25\%) \\
Improver claims with a matched improver comparison (P4 = Y) & 14 of 26 & 7 of 8 \\
\quad of which same start, earlier versus later improver & 4 of 26 & 1 of 8 \\
\bottomrule
\end{tabular}

\end{table}

\begin{table}[H]
\caption{Codes for the twenty anchor works selected by \citet{wu2026rsisurvey}. STOP and Meta$^n$ are also in the RSI frame.}
\label{tab:anchorcodes}
\centering
\small
\begin{tabular}{llccccccccc}
\toprule
Work & arXiv & Dom. & A1 & A2 & P1 & P3a & A4 & A5 & P4 & Tier \\
\midrule
STaR & 2203.14465 & V & N & P & N & N & P & N & N & E2 \\
Self-Instruct & 2212.10560 & M & N & Y & N & N & P & N & N & E2 \\
SPIN & 2401.01335 & M & N & Y & Y & N & P & N & N & E3 \\
Self-Rewarding & 2401.10020 & O & N & Y & Y & N & P & N & Y & E4 \\
Meta-Rewarding & 2407.19594 & O & N & P & Y & N & P & N & P & E3 \\
Absolute Zero & 2505.03335 & V & N & Y & Y & N & P & N & Y & E4 \\
Reflexion & 2303.11366 & V & P & P & Y & N & Y & Y & N & E2 \\
Voyager & 2305.16291 & V & Y & P & N & N & Y & Y & N & E2 \\
ExpeL & 2308.10144 & V & N & Y & N & Y & P & Y & N & E3 \\
OPRO & 2309.03409 & V & N & P & N & N & P & N & N & E2 \\
TextGrad & 2406.07496 & V & P & P & N & Y & P & N & N & E2 \\
ADAS & 2408.08435 & V & N & P & N & N & P & N & N & E2 \\
GEPA & 2507.19457 & V & N & Y & N & N & P & N & N & E2 \\
AlphaEvolve & 2506.13131 & V & Y & P & N & N & Y & Y & Y & E2 \\
Promptbreeder & 2309.16797 & V & N & Y & N & N & P & N & Y & E2 \\
SICA & 2504.15228 & V & Y & N & N & N & N & N & N & E1 \\
DGM & 2505.22954 & V & Y & P & N & N & P & N & Y & E2 \\
Hyperagents & 2603.19461 & M & N & Y & N & Y & P & N & Y & E4 \\
STOP & 2310.02304 & V & N & Y & Y & Y & P & N & Y & E4 \\
Meta$^n$ & 2608.24735 & V & P & P & N & Y & Y & N & Y & E4 \\
\bottomrule
\end{tabular}

\end{table}

\begin{table}[H]
\caption{Our codes against the protocol descriptions of \citet{wu2026rsisurvey} (their Table 7). ``Implied'' is the value of our item that their description implies; Y, P and N follow Appendix~\ref{app:codebook}. Test separation corresponds to A2, improver test to P4 and budget parity to A4.}
\label{tab:wuconc}
\centering
\small
\begin{tabular}{llp{6.3cm}cc}
\toprule
Work & Dimension & Wu et al. description & Implied & Ours \\
\midrule
Meta-Rewarding & Test separation & AlpacaEval 2; Arena-Hard; MT-Bench; separate judge prompts & Y & P \\
Promptbreeder & Test separation & Train-batch fitness; test splits (some random halves) & Y & Y \\
STOP & Test separation & 50 unseen LPN cases; 5 runs; 5 transfer tasks & Y & Y \\
ADAS & Test separation & ARC: 20 validation / 60 test tasks; 5 repeats; validation-guided search & Y & P \\
GEPA & Test separation & Train/validation/test & Y & Y \\
DGM & Test separation & full scores include search tasks; no strict holdout & P & P \\
Hyperagents & Test separation & Held-out domain tasks & Y & Y \\
Meta$^n$ & Test separation & CO-Bench, S2D, LawBench held out; other benchmarks include unsplit scores & P & P \\
Meta-Rewarding & Improver test & Actor/judge gains; later human-agreement gains fade (no frozen-judge control listed) & N & P \\
Promptbreeder & Improver test & operator ablations & P & Y \\
STOP & Improver test & Improver transfer & Y & Y \\
ADAS & Improver test & fixed meta-agent & N & N \\
GEPA & Improver test & fixed optimizer & N & N \\
DGM & Improver test & frozen-modifier/no-sampling archive controls & Y & Y \\
Hyperagents & Improver test & Frozen-meta/no-archive controls & Y & Y \\
Meta$^n$ & Improver test & depth/context/library ablations; fixed Omega & P & P \\
Meta-Rewarding & Budget parity & matched length control (no compute parity stated) & P & P \\
Promptbreeder & Budget parity & cost parity unestablished & P & P \\
STOP & Budget parity & LM/utility-call caps; seed/greedy baselines & P & P \\
ADAS & Budget parity & end-to-end cost parity unestablished & P & P \\
GEPA & Budget parity & rollouts not equal to compute parity & P & P \\
DGM & Budget parity & separate cost reports & P & P \\
Hyperagents & Budget parity & separate token estimates & P & P \\
Meta$^n$ & Budget parity & token-parity tests; parity is configuration-specific & Y & Y \\
\bottomrule
\end{tabular}

\end{table}

\end{document}